\documentclass[11pt]{article}
\usepackage[preprint]{acl}
\usepackage{times}
\usepackage{latexsym}
\usepackage[T1]{fontenc}
\usepackage[utf8]{inputenc}
\usepackage{microtype}
\usepackage{inconsolata}
\usepackage{graphicx}
\usepackage{booktabs}
\usepackage{amsmath}
\usepackage{amssymb}
\usepackage{multirow}
\usepackage{placeins}
\usepackage{tabularx}

\title{BEAR: Budgeted Evidence Allocation for Multi-Document Reasoning}

\author{
  \textbf{Lin Sun\textsuperscript{*}},
  \textbf{Linglin Zhang},
  \textbf{Jingang Huang},
  \textbf{Change Jia},
  \textbf{Zhengwei Cheng},
  \textbf{Xiangzheng Zhang}
\\
\\
  Qiyuan Tech
\\
  \small{
    \textsuperscript{*}\textbf{Correspondence:}
    \href{mailto:sunlin1@360.cn}{sunlin1@360.cn}
  }
}

\begin{document}
\maketitle

\begin{abstract}
We argue that multi-document reasoning is constrained not only by how much text a model can read, but also by how limited query-time evidence budget is allocated across documents and semantic granularities. Full-context inference exposes the model to broad evidence non-selectively and at high per-query cost, while flat chunk retrieval often returns locally relevant passages that are weakly organized for cross-document synthesis. We present \textbf{BEAR}, a framework for structured evidence allocation that builds hierarchical semantic indices offline and performs coarse-to-fine evidence access at query time through complementary \emph{exploration} and \emph{recovery} paths. This coarse-to-fine design can be viewed as structured evidence allocation under a fixed evidence-context budget. Across synthetic and real-world benchmarks, BEAR performs particularly strongly on DragonBall, remains competitive with strong retrieval-based baselines on HotpotQA, and yields the best retrieval-based result on 2Wiki under our evaluated protocol, while operating under substantially smaller \emph{query-time evidence budgets} than the reported long-context references. Additional analyses suggest that the gains are associated with hierarchy as an allocation substrate together with complementary exploration and recovery, rather than semantic chunking alone.
\end{abstract}

\section{Introduction}
Multi-document reasoning requires more than simply exposing an LLM to a large amount of text. In realistic settings, systems must answer under limited inference-time resources, including constraints on retrieved evidence, generator context, and query-time compute. A central challenge is therefore how limited \emph{evidence budget} should be allocated across documents and semantic granularities. Even with long context windows, relevant evidence may be dispersed across documents or buried in distracting context, while flat retrieval often returns passages that are locally similar to the query but weakly organized for cross-document synthesis~\citep{liu2023lost,kamradt2023needle,Comanici2025Gemini2P,bai2024longbench,ram2023context,shi2023replug}.

\begin{figure*}[t]
    \centering
    \includegraphics[width=0.95\linewidth]{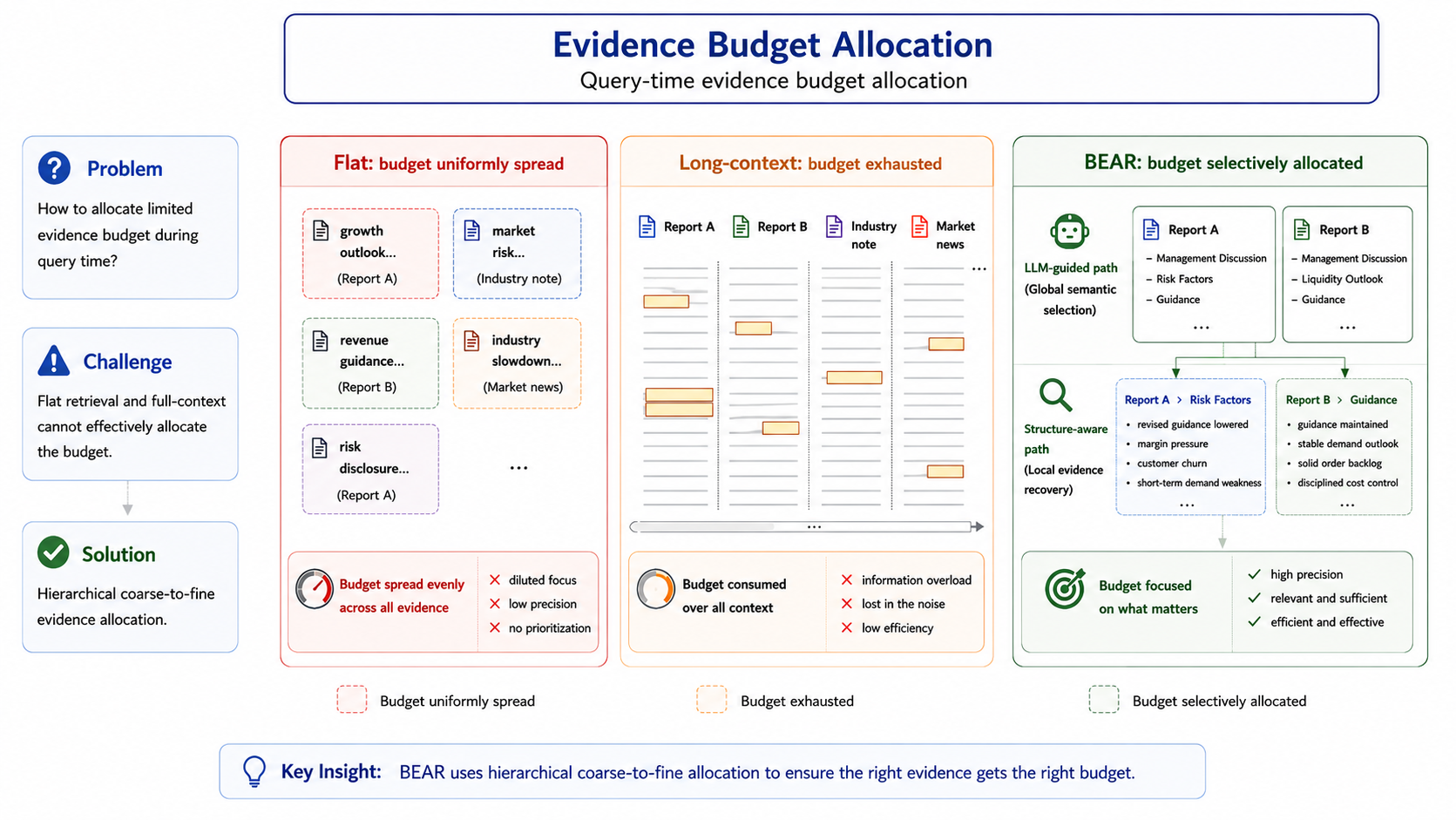}
    \caption{Motivating example for complementary exploration and recovery under a limited evidence budget.}
    \label{fig:motivation_example}
\end{figure*}

Figure~\ref{fig:motivation_example} illustrates a financial comparison query that requires both coarse evidence allocation and localized evidence recovery. Flat retrieval tends to over-retrieve weakly useful chunks, while naive long-context inference leaves evidence prioritization and localization implicit inside generation. A structured system can instead expose coarse semantic abstractions for global allocation and refine to local evidence only when needed.

\begin{figure*}[t]
    \centering
    \includegraphics[width=0.95\linewidth]{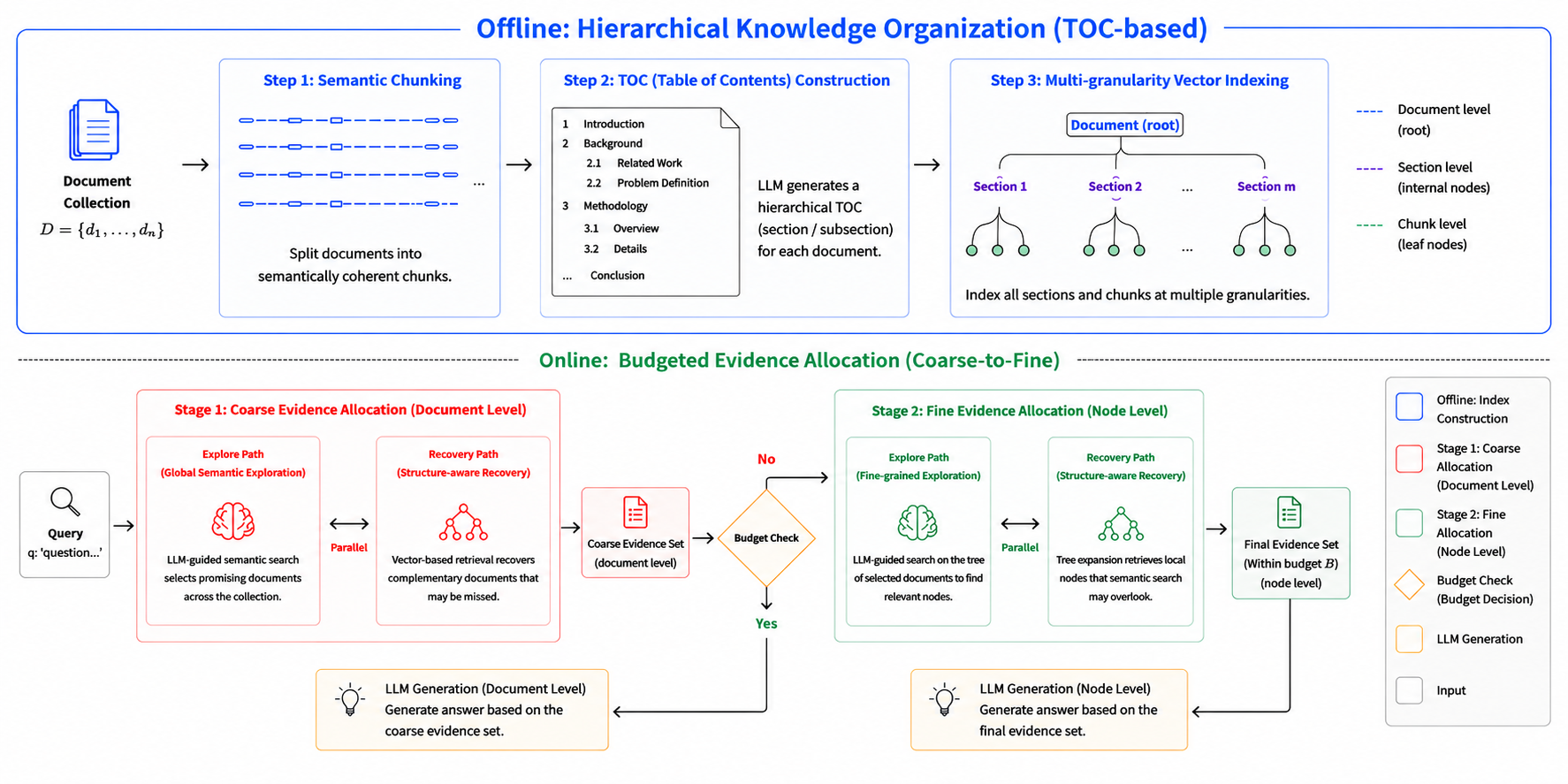}
    \caption{Overview of BEAR. Offline, it builds hierarchical semantic indices and multi-granularity representations. Online, it progressively allocates evidence through complementary exploration and recovery under a fixed evidence budget, refining to finer-grained evidence only when needed.}
    \label{fig:fable_framework}
\end{figure*}

Motivated by this, we propose \textbf{BEAR}, a framework for budget-aware evidence allocation in multi-document reasoning. BEAR organizes documents offline into hierarchical semantic trees and accesses them online through two complementary paths (Figure~\ref{fig:fable_framework}): an \emph{exploration path}, implemented via LLM-guided semantic selection over high-level summaries, and a \emph{recovery path}, implemented via TreeExpansion for localized evidence recovery. Hierarchy provides the substrate for coarse-to-fine query-time access: a small coarse-grained budget first identifies promising semantic regions, after which finer-grained evidence access focuses on detailed supporting evidence. This budget compression helps preserve global coverage without paying the full cost of exhaustive flat retrieval or long-context processing. Relative to flat retrieval, this preserves coarse semantic abstractions and multi-granularity access; relative to naive long-context inference, it makes evidence organization explicit.

BEAR formulates evidence access over hierarchical semantic indices under a fixed query-time evidence budget. This yields two operating regimes: \textbf{BEAR(docs)} for budget-efficient document-level fusion and \textbf{BEAR(nodes)} when coarse evidence exceeds the budget or finer evidence discrimination is needed. Our contributions are threefold: (1) we frame multi-document reasoning as a problem of budgeted evidence allocation across semantic granularities; (2) we introduce a coarse-to-fine evidence allocation framework with complementary exploration and recovery paths, in which hierarchy serves as a mechanism for budget-aware evidence access rather than only an indexing structure; and (3) we provide controlled evaluations across synthetic, real-world, and downstream agentic settings suggesting that this design improves over flat retrieval in our controlled ablations and offers a favorable quality--cost operating point across document-level and node-level access regimes.

\section{Related Work}
Long-context LLMs can process substantially more text, but longer context does not reliably yield stronger knowledge-intensive reasoning and increases computational cost~\citep{liu2023lost,kamradt2023needle,Comanici2025Gemini2P,bai2024longbench}. This has sustained interest in retrieval-based systems, especially when evidence must be selected rather than merely concatenated~\citep{lewis2020retrieval,gao2023retrieval}.

Structured retrieval methods improve on flat chunks by introducing graph or hierarchical organization. Graph-based systems such as GraphRAG, LightRAG, and HippoRAG organize evidence around entities and relations, while hierarchical methods such as RAPTOR and HiRAG build tree-like abstractions over documents~\citep{edge2024graphrag,guo2024lightrag,gutierrez2024hipporag,sarthi2024raptor,huang2025hirag,jin-etal-2025-hierarchical}. BEAR builds on this line of work by focusing on query-time access under a fixed evidence budget: the emphasis is not only hierarchy construction, but coarse-to-fine traversal and fusion under budget constraints. Related systems such as EfficientRAG and Tree of Reviews similarly highlight the value of efficient or iterative structured retrieval for multi-hop reasoning~\citep{zhuang2024efficientragefficientretrievermultihop,jiapeng2024treereviewstreebaseddynamic}.

A third line of work integrates LLM reasoning more directly into retrieval, including iterative retrieval-and-reasoning pipelines and agentic systems that interleave reasoning with tool use~\citep{trivedi2023interleaving,press2022measuring,asai2023selfrag,jiang2023active,yao2023react,nakano2021webgpt,schick2024toolformer}. BEAR sits between static structured retrieval and fully reactive retrieval agents: it organizes document semantics offline into hierarchical forests, then navigates that structure online through budget-aware evidence allocation and selective refinement.

\paragraph{Positioning BEAR.}
Relative to prior hierarchical or tree-based RAG, BEAR differs in emphasis: it studies how a hierarchical semantic index is used at query time under a fixed evidence budget rather than hierarchy construction in isolation. Coarse document-level evidence is preserved when sufficient, node-level refinement is invoked only when finer-grained aggregation is required, and the node-level stage combines an exploration path with a recovery path rather than a single traversal mechanism. Relative to flat retrieval and monolithic long-context inference, this design allocates a limited query-time evidence budget more selectively while making evidence organization explicit.

\section{Method}
\subsection{Overview}
BEAR uses a semantic forest (one tree per document) $\mathcal{F} = \{T_1, \dots, T_N\}$ as the substrate for budget-aware query-time access over a document collection $\mathcal{D} = \{d_1, \dots, d_N\}$. \textbf{Offline}, an LLM segments each document into coherent chunks, organizes them hierarchically, and produces summary-bearing internal nodes for high-level retrieval. Because evidence granularity itself becomes a resource-allocation decision under a fixed query-time budget, hierarchical representations provide a natural substrate for allocating limited evidence across abstraction levels. \textbf{Online}, BEAR first allocates coarse evidence candidates through a complementary exploration--recovery stage, then optionally refines them at node level if the coarse result exceeds evidence-context budget.

BEAR is designed to support both \emph{global semantic exploration} and \emph{localized evidence recovery}: one path uses LLM reasoning over hierarchical summaries to identify relevant regions, while the other uses dense similarity and structural propagation to recover supporting evidence that a purely reasoning-driven path may miss.

\subsection{Offline Hierarchical Knowledge Organization}
For each document $d_i$, BEAR first performs LLM-guided semantic chunking instead of fixed-length splitting, preserving discourse coherence and aligning chunk boundaries with semantic units. Let
\begin{equation}
\mathcal{C}_i = \mathrm{LLM}_{\mathrm{segment}}(d_i) = \{c_1, c_2, \dots, c_{m_i}\}
\end{equation}
denote the resulting chunk sequence.

BEAR then constructs a semantic tree
\begin{equation}
T_i = \mathrm{LLM}_{\mathrm{structure}}(\mathcal{C}_i \mid d_i),
\end{equation}
whose internal nodes provide topic-level abstractions (e.g., titles and summaries) for high-level retrieval, while the leaves retain the underlying chunk content needed for answer generation.

To support multi-granularity retrieval, BEAR builds separate embeddings for internal and leaf nodes. For a non-leaf node $v$, the embedding is computed from the path-level table-of-contents signal together with its summary:
\begin{equation}
\mathbf{e}_v = \mathrm{Embed}(\mathrm{toc\_path}(v) \oplus \mathrm{summary}(v)).
\end{equation}
For a leaf node $c$, we embed its original chunk content:
\begin{equation}
\mathbf{e}_c = \mathrm{Embed}(\mathrm{content}(c)).
\end{equation}
These vectors form the basis of the dense retrieval path at both document and node levels.

\subsection{Budget-Aware Evidence Allocation}
\paragraph{Coarse evidence access.}
At query time, BEAR first performs coarse evidence access through two complementary paths. The first, the \emph{exploration path}, exposes only high-level non-leaf summaries to the LLM, allowing it to reason over compact semantic abstractions. The second, the \emph{recovery path}, performs dense retrieval over the indexed nodes. Their union forms a deduplicated candidate set $\mathcal{D}_{\mathrm{fusion}}$, combining semantic reasoning with embedding-space coverage.

\paragraph{Budget-aware coarse-to-fine routing.}
Here, the query-time budget reflects the limited evidence bandwidth available to expose supporting evidence before generation. In practice, we instantiate this constraint using a token budget over the final retrieved evidence context, denoted by $B_{\max}$. BEAR first checks whether the fused document-level evidence fits within this budget. If so, BEAR directly returns the selected evidence context; otherwise, BEAR proceeds to a finer node-level stage. In the experiments, \textbf{BEAR(docs)} returns the fused document-level context, while \textbf{BEAR(nodes)} performs additional node-level exploration--recovery refinement when coarse evidence exceeds the budget or finer evidence discrimination is needed.

\paragraph{Fine-grained evidence refinement.}
When refinement is needed, BEAR again uses two paths. The first, the exploration path, asks the LLM to navigate hierarchical summaries within the selected documents and identify promising semantic nodes. The second, the recovery path, applies TreeExpansion, a lightweight tree-restricted evidence recovery mechanism that supplements direct query-node similarity with signals inherited from ancestors and aggregated from children. For a node $v$, we score it as:
\begin{equation}
S(v)=\alpha S_{\mathrm{sim}}(v)+\beta S_{\mathrm{inh}}(v)+\gamma S_{\mathrm{child}}(v),
\end{equation}
where $S_{\mathrm{sim}}(v)$ is direct query--node similarity, $S_{\mathrm{inh}}(v)$ propagates relevance from ancestors, and $S_{\mathrm{child}}(v)$ aggregates support from descendants. Unless otherwise specified, we use $\alpha=\beta=\gamma=\frac{1}{3}$ as a simple untuned default that does not require query-type prediction or additional online adaptation. Appendix Table~\ref{tab:sensitivity} shows that some query categories prefer different weightings, but we keep the equal-weight setting to preserve a fixed and lightweight retrieval policy. This scoring rule complements summary-level semantic guidance under the same evidence-budget constraint. The final evidence list is deduplicated and truncated to satisfy the budget.


\begin{table*}[t]
\centering
\caption{Results on DragonBall, HotpotQA, and 2Wiki (mean $\pm$ standard deviation). \textbf{BEAR(docs)} returns fused document-level context; \textbf{BEAR(nodes)} adds node-level refinement. Retrieval-based methods share the same Qwen3-32B generator and, where applicable, DeepSeek-V3.2 for offline and retrieval-time steps; LongRefiner is the main exception. For DragonBall, we report Recall, Completeness, Hallucination, and Irrelevance; Recall is omitted for Gemini because these references receive full source documents rather than retrieved evidence.}
\label{tab:main-results}
\small
\resizebox{\textwidth}{!}{%
\begin{tabular}{lcccccccc}
\toprule
& \multicolumn{4}{c}{Synthetic Knowledge} & \multicolumn{4}{c}{Real-World Knowledge} \\
\cmidrule(lr){2-5} \cmidrule(lr){6-9}
Method & \multicolumn{4}{c}{DragonBall} & \multicolumn{2}{c}{HotpotQA} & \multicolumn{2}{c}{2Wiki}\\
\cmidrule(lr){2-5} \cmidrule(lr){6-7} \cmidrule(lr){8-9}
& Recall(\%) $\uparrow$ & Comp.(\%) $\uparrow$ & Hall.(\%) $\downarrow$ & Irr.(\%) $\downarrow$ & EM(\%) $\uparrow$ & F1(\%) $\uparrow$ & EM(\%) $\uparrow$ & F1(\%) $\uparrow$ \\
\midrule
\multicolumn{9}{l}{\textbf{Matched Retrieval Baselines}} \\
\midrule
BM25 & 29.17 $\pm$ 1.06 & 35.13 $\pm$ 1.84 & 32.55 $\pm$ 2.40 & 32.25 $\pm$ 1.95 & 20.67 $\pm$ 0.76 & 32.69 $\pm$ 0.53 & 10.33 $\pm$ 1.44 & 18.12 $\pm$ 1.32 \\
BGE-M3 & 58.17 $\pm$ 1.91 & 59.56 $\pm$ 2.45 & 28.81 $\pm$ 2.26 & 11.54 $\pm$ 1.55 & 22.83 $\pm$ 2.08 & 36.25 $\pm$ 2.83 & 15.83 $\pm$ 1.26 & 26.94 $\pm$ 1.17 \\
RAPTOR & 31.93 $\pm$ 1.66 & 59.78 $\pm$ 2.70 & 26.50 $\pm$ 2.48 & 13.67 $\pm$ 1.29 & 29.33 $\pm$ 1.26 & 44.72 $\pm$ 1.70 & 23.50 $\pm$ 1.00 & 36.96 $\pm$ 1.30 \\
LongRefiner & 29.24 $\pm$ 2.32 & 43.86 $\pm$ 1.98 & 34.88 $\pm$ 2.25 & 21.26 $\pm$ 1.87 & 22.33 $\pm$ 0.85 & 33.97 $\pm$ 0.71 & 9.00 $\pm$ 0.87 & 13.16 $\pm$ 0.83 \\
HippoRAG2 & 68.80 $\pm$ 1.75 & 73.37 $\pm$ 2.24 & 19.55 $\pm$ 1.84 & 6.93 $\pm$ 0.93 & \textbf{33.83 $\pm$ 0.58} & \textbf{48.99 $\pm$ 1.22} & 29.67 $\pm$ 0.76 & 44.19 $\pm$ 0.73 \\
\midrule
\multicolumn{9}{l}{\textbf{Full-context Controls / References}} \\
\midrule
Gemini-2.5-Flash & -- & 88.62 $\pm$ 0.34 & 6.30 $\pm$ 0.14 & 5.09 $\pm$ 0.04 & 29.83 $\pm$ 1.31 & 45.96 $\pm$ 1.06 & 38.67 $\pm$ 0.24 & 55.21 $\pm$ 0.27 \\
Gemini-2.5-Pro & -- & 90.86 $\pm$ 0.07 & 5.46 $\pm$ 0.24 & 3.67 $\pm$ 0.09 & 41.74 $\pm$ 0.33 & 56.66 $\pm$ 0.90 & 53.01 $\pm$ 1.41 & 64.69 $\pm$ 0.72 \\
\midrule
\multicolumn{9}{l}{\textbf{Our Proposed Method}} \\
\midrule
\textbf{BEAR(docs)} & \textbf{85.55 $\pm$ 0.47} & \textbf{91.86 $\pm$ 1.32} & \textbf{5.64 $\pm$ 1.02} & \textbf{2.40 $\pm$ 0.49} & 31.00 $\pm$ 1.15 & 45.80 $\pm$ 1.18 & \textbf{34.63 $\pm$ 2.17} & \textbf{48.02 $\pm$ 1.45} \\
\textbf{BEAR(nodes)} & \underline{84.25 $\pm$ 0.84} & \underline{88.64 $\pm$ 1.52} & \underline{7.60 $\pm$ 1.12} & \underline{3.71 $\pm$ 0.86} & \underline{33.50 $\pm$ 1.15} & \underline{48.44 $\pm$ 0.54} & \underline{33.00 $\pm$ 1.73} & \underline{46.30 $\pm$ 1.48} \\
\bottomrule
\end{tabular}
}
\end{table*}

\section{Experimental Setup}
\paragraph{Datasets.}
We evaluate BEAR on synthetic and real-world multi-document reasoning benchmarks. For synthetic evaluation, we use DragonBall~\citep{zhu-etal-2025-rageval}, including settings stressing evidence integration, hallucination, and multilingual retrieval. DragonBall's use of LLM-synthesized knowledge provides a cleaner signal for retrieval quality: since modern LLMs are trained on Wikipedia, real-world benchmarks such as HotpotQA and 2Wiki risk conflating retrieval gains with parametric knowledge recall, whereas synthetic knowledge isolates retrieval as the primary variable. For real-world multi-hop reasoning, we use HotpotQA~\citep{yang2018hotpotqa} and 2Wiki~\citep{ho-etal-2020-constructing}, adapted for retrieval over candidate document collections. We additionally use BrowseComp-plus~\citep{browsecompplus} as a downstream agentic benchmark to test whether retrieval gains transfer beyond benchmark-style QA.

\paragraph{Baselines.}
We compare against sparse and dense retrieval baselines (BM25 and BGE-M3~\citep{chen-etal-2024-m3}), structure-enhanced baselines (RAPTOR, LongRefiner~\citep{jin-etal-2025-hierarchical}, and HippoRAG2~\citep{gutierrez2024hipporag,gutierrez2025ragmemory}), a same-generator full-context control (Qwen3-32B~\citep{qwen3technicalreport}), and long-context references (Gemini-2.5-Flash and Gemini-2.5-Pro). RAPTOR is included as a well-known tree-structured baseline closely related to BEAR's hierarchical indexing setting, while HippoRAG2 serves as a strong graph-based structured retriever.

For reproducibility, we use public implementations whenever available: BM25 and BGE-M3 are based on FlashRAG~\citep{Jin2024FlashRAGAM}, while RAPTOR, LongRefiner and HippoRAG2 use their official repositories. Unless a baseline has an intrinsic model dependency, offline and retrieval-time LLM components are aligned to DeepSeek-V3.2~\citep{deepseekai2025deepseekv32} and the final generator is fixed to Qwen3-32B; LongRefiner is the main exception because its preprocessing relies on a separately trained helper model. Dense retrieval is aligned to BGE-M3, with BGE-Reranker-v2-M3~\citep{li2023making} used when reranking is required. Other shared settings appear in Appendix Table~\ref{tab:default-settings-appendix}.

\begin{figure*}[t]
    \centering
    \includegraphics[width=1\linewidth]{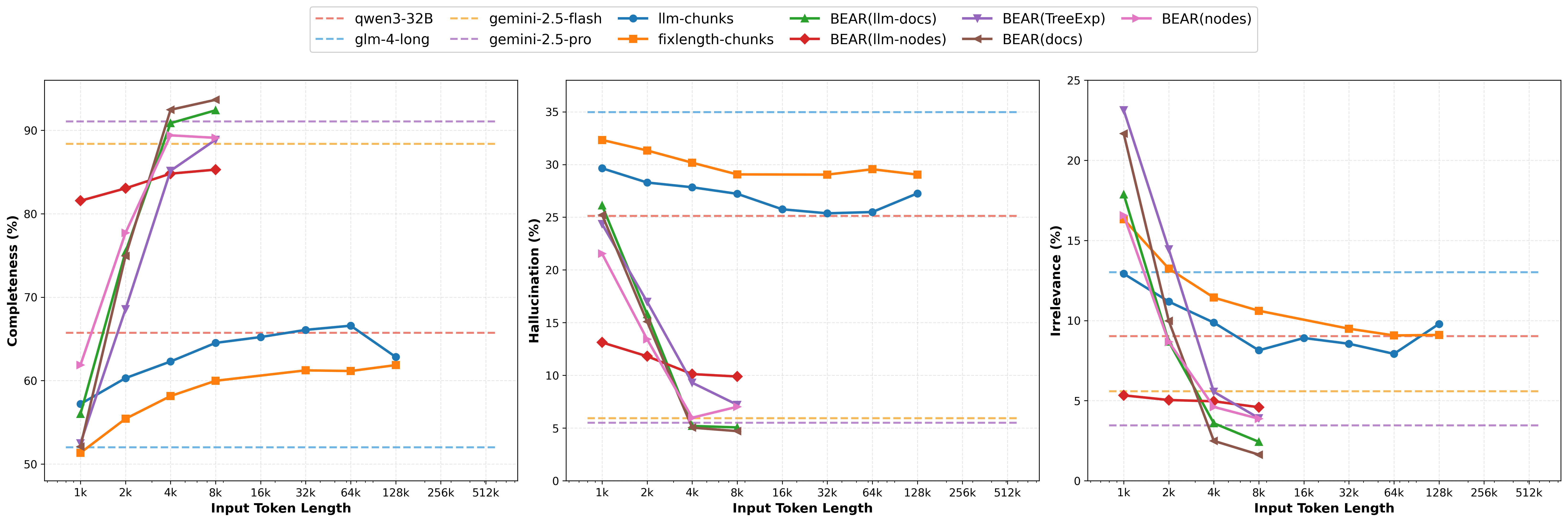}
    \caption{Performance under different evidence-context budgets. Chunk-based controls use fixed-size chunking (\textit{fixlength-chunks}) or semantic chunking (\textit{llm-chunks}) within the same Standard RAG pipeline; the remaining curves are BEAR variants. In the evaluated setting, structured evidence allocation yields a stronger completeness--faithfulness trade-off than the chunk-based controls at much smaller budgets.}
    \label{fig:comp_hall_irr}
\end{figure*}

\paragraph{Comparison protocol.}
We report four analyses: matched retrieval baselines, a same-generator full-context control, controlled ablations, and long-context references. In the matched retrieval comparisons, we keep the answer generator fixed and, where applicable, also align the offline and retrieval-time LLMs so differences are primarily attributable to retrieval policy rather than generation or auxiliary models. Qwen3-32B serves as the same-generator full-context control, while Figure~\ref{fig:comp_hall_irr} and the progressive ablations isolate the effects of semantic chunking, hierarchical indexing, and bi-path retrieval. Gemini-2.5-Flash and Gemini-2.5-Pro are contextual long-context references rather than matched retrieval baselines. Across retrieval-based methods, the final evidence context passed to the common Qwen3-32B generator is aligned to approximately 4K tokens whenever applicable, reducing confounding from unequal generator context length. When public baseline defaults yield substantially shorter generator-side contexts, we preserve the original retrieval logic and adjust only the final evidence packing or truncation to better match the shared budget. This keeps the comparison focused on retrieval behavior while controlling for unequal generator-side context length; accordingly, our main claim is about the operating point achieved under a shared-generator, budget-matched retrieval protocol rather than universal superiority across model families or serving stacks. For BrowseComp-plus, we keep the agent policy model fixed and replace only the retriever with BEAR.

\paragraph{Metrics and protocol.}
For DragonBall, we report Recall, Completeness, Hallucination, and Irrelevance, following~\citep{zhu-etal-2025-rageval}. For HotpotQA and 2Wiki, we report EM and F1 using the official HippoRAG2 scripts. Unless otherwise noted, we report mean and standard deviation over 3 repeated evaluation runs under a fixed evaluation setup. We also analyze token cost, end-to-end latency, and the contribution of bi-path retrieval; robustness of TreeExpansion appears in Appendix Table~\ref{tab:sensitivity}.

\section{Results}
We organize the results around three questions: whether BEAR improves the budget-constrained operating point, where the gains come from, and whether they persist in matched, cost, and transfer analyses.

\subsection{Main Results}
Table~\ref{tab:main-results} shows that BEAR performs particularly strongly on DragonBall among the evaluated retrieval-based methods: both variants substantially outperform the matched retrieval baselines, and \textbf{BEAR(docs)} provides the strongest retrieval-based trade-off in the table. On HotpotQA, \textbf{BEAR(nodes)} remains close to HippoRAG2, while on 2Wiki \textbf{BEAR(docs)} gives the strongest retrieval-based result under the evaluated protocol. We include Qwen3-32B as a full-context control and Gemini-2.5-Flash/Pro as long-context references rather than matched retrieval baselines.

On the synthetic benchmark, the clearest advantage is the coverage--faithfulness trade-off. On DragonBall, \textbf{BEAR(docs)} reaches 85.55\% Recall and 91.86\% Completeness while keeping Hallucination and Irrelevance at 5.64\% and 2.40\%, respectively. Table~\ref{tab:qwen-fullcontext-dragonball} further shows that, under the same Qwen3-32B generator, passing the full document set directly to the model sharply reduces answer quality, suggesting that the gains are associated with evidence organization and selection rather than with the answer model alone. On the real-world multi-hop benchmarks, the two variants reflect different operating regimes. \textbf{BEAR(nodes)} is competitive on HotpotQA and remains close to HippoRAG2; HippoRAG2's KG construction explicitly links entities across documents, giving it a structural advantage on HotpotQA's entity-centric multi-hop design, while BEAR's per-document semantic trees do not model cross-document entity links. On 2Wiki, \textbf{BEAR(docs)} performs best among the compared retrieval-based methods. \textbf{BEAR(docs)} is preferable when coarse-grained evidence allocation already yields a coherent evidence set, whereas \textbf{BEAR(nodes)} is preferable when finer-grained evidence aggregation improves precision.

Figure~\ref{fig:comp_hall_irr} further clarifies these gains under different evidence-context budgets. Semantic chunking consistently improves over fixed-size chunking, but both chunk-based variants remain well below the hierarchical BEAR variants, especially in the low- and medium-budget regimes. Under very small budgets (1K--2K), \textbf{BEAR(llm-nodes)} is the strongest node-level variant. At moderate budgets (4K--8K), the structured variants become competitive with the reported long-context references in this contextual comparison while using 4K--8K retrieved tokens rather than 512K-token full-context inputs. Together, these controls suggest that the gains are associated with hierarchy as an allocation substrate together with complementary exploration--recovery evidence allocation rather than with semantic chunking alone or simply exposing more text.

\begin{table}[t]
\centering
\caption{DragonBall full-context control under the same Qwen3-32B generator. Entries report mean $\pm$ standard deviation over repeated runs. The comparison isolates the effect of passing the full document set directly to the generator versus structured evidence selection.}
\label{tab:qwen-fullcontext-dragonball}
\small
\setlength{\tabcolsep}{4pt}
\begin{tabular}{lccc}
\toprule
Method & Comp. (\%) & Hall. (\%) & Irr. (\%) \\
\midrule
Qwen3-32B & 66.86 $\pm$ 0.30 & 25.36 $\pm$ 0.33 & 7.70 $\pm$ 0.01 \\
\textbf{BEAR(docs)} & \textbf{91.86 $\pm$ 1.32} & \textbf{5.64 $\pm$ 1.02} & \textbf{2.40 $\pm$ 0.49} \\
\textbf{BEAR(nodes)} & \underline{88.64 $\pm$ 1.52} & \underline{7.60 $\pm$ 1.12} & \underline{3.71 $\pm$ 0.86} \\
\bottomrule
\end{tabular}
\end{table}

\subsection{Why exploration--recovery helps}
We focus on the node-level stage, where the interaction between the exploration and recovery paths is most directly observable. Table~\ref{tab:querytype-bipath} shows that neither path is uniformly strongest across answerable query types: the recovery path (TreeExpansion) is stronger on factual and temporal questions (FQ, TSQ), where localized or order-sensitive evidence matters, while the exploration path (LLM-guided selection) is stronger or competitive on synthesis-heavy categories such as information integration (IIQ), numerical comparison (NCQ), multi-hop reasoning (MRQ), and summarization (SQ). Bi-path fusion improves over the stronger single path on every answerable query type, with gains from +2.08 to +8.99 points and an average of +6.01. Table~\ref{tab:bipath-overlap} shows the paths are not redundant: 80.7\% of supported evidence is recovered by both, while 10.3\% is found only by TreeExpansion and 6.1\% only by the LLM-guided path. Multi-hop Reasoning is the clearest hard case, with overlap dropping to 72.3\% and the uncovered portion rising to 8.4\%. Error analysis shows that the ``neither'' cases are concentrated in the medical subdomain and mostly involve deep leaf nodes for multi-hop or comparison queries. On the exploration path, the main failure is conclusion-node selection bias: the model often selects the semantically closest summary node whose subtree does not cover the evidence-bearing sibling branches required for the answer. On the recovery path, misses arise from entity-sparse retrieval in a homogeneous corpus or from cross-branch truncation under the fixed evidence budget; Appendix~\ref{app:neither-failure} provides the full breakdown.

The progressive ablation supports a staged attribution: semantic chunking helps, evidence access over the hierarchical index provides the largest improvement over flat retrieval in the evaluated ablation, and complementary bi-path evidence access supplies the final increase in evidence coverage. The largest jump comes from moving from leaf-only evidence access to tree-aware evidence access (50.55\%$\rightarrow$87.91\%), indicating that the hierarchical index provides a substantially stronger allocation substrate than flat retrieval built on the same semantically chunked leaves. This gain reflects the value of the LLM-constructed hierarchical semantic index, including internal summary-bearing nodes and hierarchical organization, rather than a control that isolates bare tree topology. On top of this substrate, TreeExpansion, LLM-guided evidence selection, and especially their fusion provide further gains by recovering complementary evidence under the same budget. Appendix Table~\ref{tab:llm-swap-sensitivity} further shows that swapping the offline and online LLMs mainly changes the relative strength of the single-path components, especially the online exploration path, while the final bi-path system remains comparatively stable. Full query-type results appear in Appendix Table~\ref{tab:progressive-retrieval-ablation}, and Appendix Table~\ref{tab:domain-language-bipath} shows that this complementarity remains stable across domains and languages. The magnitude of the leaf-to-tree jump also suggests a structural prerequisite: gains from hierarchical indexing are largest when document organization is rich enough for LLM-built parent summaries to be informative, and may shrink on corpora where that structure is weak or absent.

\begin{table}[t]
\centering
\caption{Fine-grained evidence analysis across query types on answerable questions. Gain reports the improvement of bi-path fusion over the stronger single path; higher is better.}
\label{tab:querytype-bipath}
\small
\setlength{\tabcolsep}{1.5pt}
\begin{tabular}{@{}lcccc@{}}
\toprule
Type & LLM-guided (\%) & TreeExp. (\%) & Bi-path (\%) & Gain \\
\midrule
FQ & 87.50 & 95.42 & \textbf{99.17} & +3.75 pp \\
IIQ & 88.39 & 90.08 & \textbf{96.68} & +6.60 pp \\
NCQ & 90.35 & 90.07 & \textbf{97.22} & +6.87 pp \\
TSQ & 89.65 & 95.42 & \textbf{97.50} & +2.08 pp \\
MRQ & 87.12 & 86.24 & \textbf{96.11} & +8.99 pp \\
SQ & 91.35 & 91.55 & \textbf{99.31} & +7.76 pp \\
\bottomrule
\end{tabular}
\end{table}

\begin{table}[t]
\centering
\caption{Progressive ablation of node-level evidence coverage. All variants use the same LLM-based semantic chunking; Leaf-only restricts evidence access to leaf chunks, while Tree-aware uses hierarchy-aware evidence access over the hierarchical index. Full query-type results appear in Appendix Table~\ref{tab:progressive-retrieval-ablation}.}
\label{tab:progressive-ablation-main}
\small
\setlength{\tabcolsep}{1.5pt}
\begin{tabular}{@{}lcccc@{}}
\toprule
Variant & Hier. & LLM-guided & TreeExp. & Recall (\%) \\
\midrule
Leaf-only         & $\times$ & $\times$ & $\times$ & 50.55 \\
Tree-aware        & \checkmark & $\times$ & $\times$ & 87.91 \\
Tree + LLM-guided & \checkmark & \checkmark & $\times$ & 89.25 \\
Tree + TreeExp.   & \checkmark & $\times$ & \checkmark & 91.55 \\
\textbf{BEAR(nodes)} & \checkmark & \checkmark & \checkmark & \textbf{98.19} \\
\bottomrule
\end{tabular}
\end{table}

\begin{table*}[t]
\centering
\caption{Online serving token usage and latency. Retrieval-based systems share the same Qwen3-32B generator; Gemini is a long-context reference. Token counts are in thousands (K) and averaged per query.}
\label{tab:online-cost-latency}
\small
\begin{tabular}{lccccccc}
\toprule
& \multicolumn{2}{c}{Retrieval (K tokens)} & \multicolumn{2}{c}{Generation (K tokens)} & \multicolumn{3}{c}{Latency (s)} \\
\cmidrule(lr){2-3} \cmidrule(lr){4-5} \cmidrule(lr){6-8}
Method & In & Out & In & Out & Retrieval & Generation & Total\\
\midrule
\textbf{BEAR(nodes)} & 27.95 & \underline{0.02} & \textbf{3.52} & \textbf{0.14} & 6.81 & \textbf{16.71} & \textbf{23.52} \\
Standard RAG & \textbf{0} & \textbf{0} & 4.21 & 0.89 & \textbf{0.11} & 27.37 & 27.48 \\
HippoRAG2 & \underline{2.91} & 0.04 & 4.21 & 0.79 & 7.15 & 25.42 & 32.57 \\
LongRefiner & 129.25 & 0.59 & 3.94 & \underline{0.63} & 17.16 & \underline{20.90} & 38.06 \\
RAPTOR & \textbf{0} & \textbf{0} & \underline{3.79} & 1.01 & \underline{5.47} & 21.40 & \underline{26.87} \\
\midrule
Gemini-2.5-Flash & 0 & 0 & 397.75 & 0.36 & 0 & 19.23 & 19.23 \\
Gemini-2.5-Pro & 0 & 0 & 397.75 & 0.80 & 0 & 19.34 & 19.34 \\
\bottomrule
\end{tabular}
\end{table*}

\subsection{Transfer to agentic retrieval}
We additionally evaluate BEAR on BrowseComp-plus, an agentic search-and-synthesis benchmark over large document collections. With the agent policy model fixed, replacing only the retriever with BEAR improves accuracy from 44.46\% to 66.60\% and recall from 62.32\% to 76.60\%, suggesting that the retrieval gains can transfer beyond benchmark-style QA in this controlled agent replacement setting. Appendix Table~\ref{tab:main-results-agent} also shows that BEAR reduces the number of search calls relative to the original retriever only in the reported Tongyi-DeepResearch comparison setup.

\subsection{Cost and Latency Analysis}
We separate one-time indexing workload from online serving cost because systems distribute preprocessing and serving effort differently.

\paragraph{Online serving cost.}
Table~\ref{tab:online-cost-latency} reports online token usage and latency under the same budget-matched protocol as the main retrieval comparisons, so quality and efficiency are evaluated under the same setting. Within the retrieval-based group, all final answers are generated by the same Qwen3-32B backbone. Under this controlled protocol, BEAR provides a favorable online quality--latency profile among the retrieval-based methods we evaluate. Relative to LongRefiner and HippoRAG2, it substantially reduces total latency, and relative to Standard RAG and RAPTOR, it shifts more computation into retrieval while still achieving lower end-to-end latency. Overall, these results suggest that BEAR pays a moderate retrieval-stage cost for a stronger end-to-end quality--latency trade-off than the alternative retrieval-based baselines we compare.

We also report Gemini-2.5-Flash and Gemini-2.5-Pro as long-context references. Their online burden falls almost entirely on generation, reaching 397.75K input tokens on average, and their latency values are contextual because they rely on different long-context models and serving stacks. These measurements complement, rather than redefine, BEAR's main query-time budget: the token budget on retrieved evidence before final generation.

\paragraph{Offline indexing workload.}
Offline preprocessing cost is reported as token volume because the indexing pipelines rely on different auxiliary models. BEAR, RAPTOR and HippoRAG2 use DeepSeek-V3.2 for LLM-assisted index construction, whereas LongRefiner relies on a separately trained Qwen2.5-3B-Instruct helper model. Under this view, BEAR requires substantially fewer offline input tokens than HippoRAG2 (0.95M vs.\ 3.74M), but more than RAPTOR (0.46M) and LongRefiner (0.39M). At the same time, BEAR produces substantially more output tokens because it builds a summary-rich hierarchical semantic index rather than a lighter preprocessing artifact. Most of BEAR's offline workload comes from semantic chunking and tree construction. Maintaining the hierarchy under document updates introduces additional systems overhead not captured by token volume alone.

\begin{table}[t]
\centering
\caption{Offline indexing token workload. Token counts are in millions (M). LongRefiner uses a separate Qwen2.5-3B-Instruct helper model in preprocessing.}
\label{tab:offline-cost}
\small
\begin{tabular}{lcc}
\toprule
Method / stage & Input (M) & Output (M) \\
\midrule
\textbf{BEAR(nodes)} & 0.95 & 2.65 \\
\quad chunking & 0.43 & 1.48 \\
\quad tree building & 0.53 & 1.17 \\
\midrule
HippoRAG2 & 3.74 & 0.88 \\
LongRefiner & 0.39 & 0.22 \\
RAPTOR & 0.46 & 0.09 \\
\bottomrule
\end{tabular}
\end{table}

\section{Conclusion}
We presented \textbf{BEAR}, a framework for budgeted evidence allocation in multi-document reasoning that combines hierarchical semantic indexing, complementary exploration--recovery evidence allocation, and budget-aware query-time access under a fixed evidence-context budget. Across the evaluated benchmarks, BEAR performs particularly strongly on DragonBall, remains competitive on HotpotQA, and yields the best retrieval-based result on 2Wiki under our matched protocol, while using smaller \emph{query-time evidence budgets} than the reported long-context references. Coarse-grained evidence access provides a favorable completeness--faithfulness trade-off, while node-level refinement is most useful when finer-grained evidence aggregation is needed. Overall, our findings suggest that increasing available context alone is insufficient; evidence exposure itself must be selectively organized. BEAR occupies a favorable operating regime for semantically organized corpora under constrained query-time budgets, with tradeoffs in offline indexing cost, online latency, and dependence on semantically structured corpora.

\clearpage

\section*{Limitations}
BEAR is most effective when the document collection has enough semantic structure for hierarchical organization to be informative. On highly noisy, weakly structured, or rapidly changing corpora, the hierarchy may become less reliable as an allocation substrate, and the gains from selective evidence allocation may diminish. The method also incurs nontrivial offline indexing cost and additional system complexity relative to standard flat RAG pipelines, since it requires semantic chunking, hierarchical tree construction, and multi-granularity indexing before query-time retrieval. In dynamic settings, practical deployment may require subtree reconstruction, index maintenance under document drift, and preserving summary consistency across abstraction levels. In our current implementation, updates are localized at the document level: when a document changes, we rerun semantic chunking, tree construction, node embedding, and the corresponding vector index updates only for that document, while leaving unchanged documents untouched. Finally, BEAR improves evidence selection and organization, but it does not by itself eliminate downstream reasoning errors of the generator. In practice, BEAR is most attractive when documents have reusable semantic organization, evidence must be selected under a fixed query-time evidence budget, and the application can tolerate offline preprocessing and structure maintenance. These tradeoffs make BEAR particularly suitable when improved budgeted reasoning quality justifies preprocessing and structured indexing, and less suitable for fully general-purpose retrieval settings. Our experiments use public benchmarks and public implementations when available, but exact reproduction may still depend on proprietary APIs and evolving serving behavior of commercial models.

\bibliography{custom}

@article{Comanici2025Gemini2P,
  title={Gemini 2.5: Pushing the frontier with advanced reasoning, multimodality, long context, and next generation agentic capabilities},
  author={Comanici, Gheorghe and Bieber, Eric and Schaekermann, Mike and Pasupat, Ice and Sachdeva, Noveen and Dhillon, Inderjit and Blistein, Marcel and Ram, Ori and Zhang, Dan and Rosen, Evan and others},
  journal={arXiv preprint arXiv:2507.06261},
  year={2025}
}

@article{liu2023lost,
    title = "Lost in the Middle: How Language Models Use Long Contexts",
    author = "Liu, Nelson F.  and
      Lin, Kevin  and
      Hewitt, John  and
      Paranjape, Ashwin  and
      Bevilacqua, Michele  and
      Petroni, Fabio  and
      Liang, Percy",
    journal = "Transactions of the Association for Computational Linguistics",
    volume = "12",
    year = "2024",
    address = "Cambridge, MA",
    publisher = "MIT Press",
    url = "https://aclanthology.org/2024.tacl-1.9/",
    doi = "10.1162/tacl_a_00638",
    pages = "157--173"
}

@misc{kamradt2023needle,
  author = {Kamradt, Greg},
  title = {Needle In A Haystack - Pressure Testing {LLMs}},
  year = {2023},
  howpublished = {\url{https://github.com/gkamradt/LLMTest_NeedleInAHaystack}},
  note = {Accessed: 2024-01-16}
}

@inproceedings{lewis2020retrieval,
  title = {Retrieval-Augmented Generation for Knowledge-Intensive {NLP} Tasks},
  author = {Lewis, Patrick and Perez, Ethan and Piktus, Aleksandra and Petroni, Fabio and Karpukhin, Vladimir and Goyal, Naman and K{\"u}ttler, Heinrich and Lewis, Mike and Yih, Wen-tau and Rockt{\"a}schel, Tim and Riedel, Sebastian and Kiela, Douwe},
  booktitle = {Advances in Neural Information Processing Systems},
  volume = {33},
  pages = {9459--9474},
  year = {2020}
}

@misc{gao2023retrieval,
      title={Retrieval-Augmented Generation for Large Language Models: A Survey}, 
      author={Yunfan Gao and Yun Xiong and Xinyu Gao and Kangxiang Jia and Jinliu Pan and Yuxi Bi and Yi Dai and Jiawei Sun and Meng Wang and Haofen Wang},
      year={2024},
      eprint={2312.10997},
      archivePrefix={arXiv},
      primaryClass={cs.CL},
      url={https://arxiv.org/abs/2312.10997}, 
}

@inproceedings{ram2023context,
  title = {In-Context Retrieval-Augmented Language Models},
  author = {Ram, Ori and Levine, Yoav and Dalmedigos, Itay and Muhlgay, Dor and Shashua, Amnon and Leyton-Brown, Kevin and Shoham, Yoav},
  booktitle = {Transactions of the Association for Computational Linguistics},
  volume = {11},
  pages = {1316--1331},
  year = {2023}
}

@article{shi2023replug,
  title = {{REPLUG}: Retrieval-Augmented Black-Box Language Models},
  author = {Shi, Weijia and Min, Sewon and Yasunaga, Michihiro and Seo, Minjoon and James, Rich and Lewis, Mike and Zettlemoyer, Luke and Yih, Wen-tau},
  journal = {arXiv preprint arXiv:2301.12652},
  year = {2023}
}

@inproceedings{trivedi2023interleaving,
  title = {Interleaving Retrieval with Chain-of-Thought Reasoning for Knowledge-Intensive Multi-Step Questions},
  author = {Trivedi, Harsh and Balasubramanian, Niranjan and Khot, Tushar and Sabharwal, Ashish},
  booktitle = {Proceedings of the 61st Annual Meeting of the Association for Computational Linguistics},
  pages = {10014--10037},
  year = {2023}
}

@inproceedings{yang2018hotpotqa,
  title = {{HotpotQA}: A Dataset for Diverse, Explainable Multi-hop Question Answering},
  author = {Yang, Zhilin and Qi, Peng and Zhang, Saizheng and Bengio, Yoshua and Cohen, William W and Salakhutdinov, Ruslan and Manning, Christopher D},
  booktitle = {Proceedings of the 2018 Conference on Empirical Methods in Natural Language Processing},
  pages = {2369--2380},
  year = {2018}
}

@article{edge2024graphrag,
  title = {From Local to Global: A Graph {RAG} Approach to Query-Focused Summarization},
  author = {Edge, Darren and Trinh, Ha and Cheng, Newman and Bradley, Joshua and Chao, Alex and Mody, Apurva and Truitt, Steven and Larson, Jonathan},
  journal = {arXiv preprint arXiv:2404.16130},
  year = {2024}
}

@article{guo2024lightrag,
  title = {{LightRAG}: Simple and Fast Retrieval-Augmented Generation},
  author = {Guo, Zirui and Lian, Xiaohua and Yang, Yanhua and Huang, Hanzhi and Liu, Shuwen and Feng, Yixuan and Liu, Yiding and Li, Jinhao},
  journal = {arXiv preprint arXiv:2410.05779},
  year = {2024}
}

@inproceedings{gutierrez2024hipporag,
  title = {{HippoRAG}: Neurobiologically Inspired Long-Term Memory for Large Language Models},
  author = {Guti{\'e}rrez, Bernal Jim{\'e}nez and Shu, Yiheng and Gu, Yu and Yasunaga, Michihiro and Su, Yu},
  booktitle = {Advances in Neural Information Processing Systems},
  volume = {37},
  year = {2024}
}

@inproceedings{sarthi2024raptor,
  title = {{RAPTOR}: Recursive Abstractive Processing for Tree-Organized Retrieval},
  author = {Sarthi, Parth and Abdullah, Salman and Tuli, Aditi and Khanna, Shubh and Goldie, Anna and Manning, Christopher D.},
  booktitle = {International Conference on Learning Representations},
  year = {2024}
}

@article{huang2025hirag,
  title = {Retrieval-Augmented Generation with Hierarchical Knowledge},
  author = {Huang, Haoyu and Huang, Yongfeng and Yang, Junjie and Pan, Zhenyu and Chen, Yongqiang and Ma, Kaili and Chen, Hongzhi and Cheng, James},
  journal = {arXiv preprint arXiv:2503.10150},
  year = {2025}
}

@misc{asai2023selfrag,
      title={Self-RAG: Learning to Retrieve, Generate, and Critique through Self-Reflection}, 
      author={Akari Asai and Zeqiu Wu and Yizhong Wang and Avirup Sil and Hannaneh Hajishirzi},
      year={2023},
      eprint={2310.11511},
      archivePrefix={arXiv},
      primaryClass={cs.CL},
      url={https://arxiv.org/abs/2310.11511}, 
}

@misc{browsecompplus,
      title={BrowseComp-Plus: A More Fair and Transparent Evaluation Benchmark of Deep-Research Agent}, 
      author={Zijian Chen and Xueguang Ma and Shengyao Zhuang and Ping Nie and Kai Zou and Andrew Liu and Joshua Green and Kshama Patel and Ruoxi Meng and Mingyi Su and Sahel Sharifymoghaddam and Yanxi Li and Haoran Hong and Xinyu Shi and Xuye Liu and Nandan Thakur and Crystina Zhang and Luyu Gao and Wenhu Chen and Jimmy Lin},
      year={2025},
      eprint={2508.06600},
      archivePrefix={arXiv},
      primaryClass={cs.CL},
      url={https://arxiv.org/abs/2508.06600}, 
}

@inproceedings{jin-etal-2025-hierarchical,
    title = "Hierarchical Document Refinement for Long-context Retrieval-augmented Generation",
    author = "Jin, Jiajie  and
      Li, Xiaoxi  and
      Dong, Guanting  and
      Zhang, Yuyao  and
      Zhu, Yutao  and
      Wu, Yongkang  and
      Li, Zhonghua  and
      Qi, Ye  and
      Dou, Zhicheng",
    editor = "Che, Wanxiang  and
      Nabende, Joyce  and
      Shutova, Ekaterina  and
      Pilehvar, Mohammad Taher",
    booktitle = "Proceedings of the 63rd Annual Meeting of the Association for Computational Linguistics (Volume 1: Long Papers)",
    month = jul,
    year = "2025",
    address = "Vienna, Austria",
    publisher = "Association for Computational Linguistics",
    url = "https://aclanthology.org/2025.acl-long.176/",
    doi = "10.18653/v1/2025.acl-long.176",
    pages = "3502--3520",
    ISBN = "979-8-89176-251-0"
}

@misc{gutierrez2025ragmemory,
      title={From RAG to Memory: Non-Parametric Continual Learning for Large Language Models},
      author={Bernal Jiménez Gutiérrez and Yiheng Shu and Weijian Qi and Sizhe Zhou and Yu Su},
      year={2025},
      eprint={2502.14802},
      archivePrefix={arXiv},
      primaryClass={cs.CL},
      url={https://arxiv.org/abs/2502.14802},
}

@inproceedings{chen-etal-2024-m3,
    title = "{M}3-Embedding: Multi-Linguality, Multi-Functionality, Multi-Granularity Text Embeddings Through Self-Knowledge Distillation",
    author = "Chen, Jianlyu  and
      Xiao, Shitao  and
      Zhang, Peitian  and
      Luo, Kun  and
      Lian, Defu  and
      Liu, Zheng",
    editor = "Ku, Lun-Wei  and
      Martins, Andre  and
      Srikumar, Vivek",
    booktitle = "Findings of the Association for Computational Linguistics: ACL 2024",
    month = aug,
    year = "2024",
    address = "Bangkok, Thailand",
    publisher = "Association for Computational Linguistics",
    url = "https://aclanthology.org/2024.findings-acl.137/",
    doi = "10.18653/v1/2024.findings-acl.137",
    pages = "2318--2335",
}

@inproceedings{zhu-etal-2025-rageval,
    title = "{RAGE}val: Scenario Specific {RAG} Evaluation Dataset Generation Framework",
    author = "Zhu, Kunlun  and
      Luo, Yifan  and
      Xu, Dingling  and
      Yan, Yukun  and
      Liu, Zhenghao  and
      Yu, Shi  and
      Wang, Ruobing  and
      Wang, Shuo  and
      Li, Yishan  and
      Zhang, Nan  and
      Han, Xu  and
      Liu, Zhiyuan  and
      Sun, Maosong",
    editor = "Che, Wanxiang  and
      Nabende, Joyce  and
      Shutova, Ekaterina  and
      Pilehvar, Mohammad Taher",
    booktitle = "Proceedings of the 63rd Annual Meeting of the Association for Computational Linguistics (Volume 1: Long Papers)",
    month = jul,
    year = "2025",
    address = "Vienna, Austria",
    publisher = "Association for Computational Linguistics",
    url = "https://aclanthology.org/2025.acl-long.418/",
    doi = "10.18653/v1/2025.acl-long.418",
    pages = "8520--8544",
    ISBN = "979-8-89176-251-0"
}

@inproceedings{ho-etal-2020-constructing,
    title = "Constructing A Multi-hop {QA} Dataset for Comprehensive Evaluation of Reasoning Steps",
    author = "Ho, Xanh  and
      Duong Nguyen, Anh-Khoa  and
      Sugawara, Saku  and
      Aizawa, Akiko",
    editor = "Scott, Donia  and
      Bel, Nuria  and
      Zong, Chengqing",
    booktitle = "Proceedings of the 28th International Conference on Computational Linguistics",
    month = dec,
    year = "2020",
    address = "Barcelona, Spain (Online)",
    publisher = "International Committee on Computational Linguistics",
    url = "https://aclanthology.org/2020.coling-main.580/",
    doi = "10.18653/v1/2020.coling-main.580",
    pages = "6609--6625"
}

@inproceedings{bai2024longbench,
  title={Longbench: A bilingual, multitask benchmark for long context understanding},
  author={Bai, Yushi and Lv, Xin and Zhang, Jiajie and Lyu, Hongchang and Tang, Jiankai and Huang, Zhidian and Du, Zhengxiao and Liu, Xiao and Zeng, Aohan and Hou, Lei and others},
  booktitle={Proceedings of the 62nd annual meeting of the association for computational linguistics (volume 1: Long papers)},
  pages={3119--3137},
  year={2024}
}

@misc{yao2023react,
      title={ReAct: Synergizing Reasoning and Acting in Language Models}, 
      author={Shunyu Yao and Jeffrey Zhao and Dian Yu and Nan Du and Izhak Shafran and Karthik Narasimhan and Yuan Cao},
      year={2023},
      eprint={2210.03629},
      archivePrefix={arXiv},
      primaryClass={cs.CL},
      url={https://arxiv.org/abs/2210.03629}, 
}

@misc{press2022measuring,
      title={Measuring and Narrowing the Compositionality Gap in Language Models}, 
      author={Ofir Press and Muru Zhang and Sewon Min and Ludwig Schmidt and Noah A. Smith and Mike Lewis},
      year={2023},
      eprint={2210.03350},
      archivePrefix={arXiv},
      primaryClass={cs.CL},
      url={https://arxiv.org/abs/2210.03350}, 
}

@misc{schick2024toolformer,
      title={Toolformer: Language Models Can Teach Themselves to Use Tools}, 
      author={Timo Schick and Jane Dwivedi-Yu and Roberto Dessì and Roberta Raileanu and Maria Lomeli and Luke Zettlemoyer and Nicola Cancedda and Thomas Scialom},
      year={2023},
      eprint={2302.04761},
      archivePrefix={arXiv},
      primaryClass={cs.CL},
      url={https://arxiv.org/abs/2302.04761}, 
}

@misc{nakano2021webgpt,
      title={WebGPT: Browser-assisted question-answering with human feedback}, 
      author={Reiichiro Nakano and Jacob Hilton and Suchir Balaji and Jeff Wu and Long Ouyang and Christina Kim and Christopher Hesse and Shantanu Jain and Vineet Kosaraju and William Saunders and Xu Jiang and Karl Cobbe and Tyna Eloundou and Gretchen Krueger and Kevin Button and Matthew Knight and Benjamin Chess and John Schulman},
      year={2022},
      eprint={2112.09332},
      archivePrefix={arXiv},
      primaryClass={cs.CL},
      url={https://arxiv.org/abs/2112.09332}, 
}

@inproceedings{jiang2023active,
    title = "Active Retrieval Augmented Generation",
    author = "Jiang, Zhengbao  and
      Xu, Frank  and
      Gao, Luyu  and
      Sun, Zhiqing  and
      Liu, Qian  and
      Dwivedi-Yu, Jane  and
      Yang, Yiming  and
      Callan, Jamie  and
      Neubig, Graham",
    editor = "Bouamor, Houda  and
      Pino, Juan  and
      Bali, Kalika",
    booktitle = "Proceedings of the 2023 Conference on Empirical Methods in Natural Language Processing",
    month = dec,
    year = "2023",
    address = "Singapore",
    publisher = "Association for Computational Linguistics",
    url = "https://aclanthology.org/2023.emnlp-main.495/",
    doi = "10.18653/v1/2023.emnlp-main.495",
    pages = "7969--7992"
}

@misc{li2023making,
      title={Making Large Language Models A Better Foundation For Dense Retrieval}, 
      author={Chaofan Li and Zheng Liu and Shitao Xiao and Yingxia Shao},
      year={2023},
      eprint={2312.15503},
      archivePrefix={arXiv},
      primaryClass={cs.CL}
}

@misc{deepseekai2025deepseekv32,
      title={DeepSeek-V3.2: Pushing the Frontier of Open Large Language Models}, 
      author={DeepSeek-AI and Aixin Liu and Aoxue Mei and Bangcai Lin and Bing Xue and Bingxuan Wang and Bingzheng Xu and Bochao Wu and Bowei Zhang and Chaofan Lin and Chen Dong and Chengda Lu and Chenggang Zhao and Chengqi Deng and Chenhao Xu and Chong Ruan and Damai Dai and Daya Guo and Dejian Yang and Deli Chen and Erhang Li and Fangqi Zhou and Fangyun Lin and Fucong Dai and Guangbo Hao and Guanting Chen and Guowei Li and H. Zhang and Hanwei Xu and Hao Li and Haofen Liang and Haoran Wei and Haowei Zhang and Haowen Luo and Haozhe Ji and Honghui Ding and Hongxuan Tang and Huanqi Cao and Huazuo Gao and Hui Qu and Hui Zeng and Jialiang Huang and Jiashi Li and Jiaxin Xu and Jiewen Hu and Jingchang Chen and Jingting Xiang and Jingyang Yuan and Jingyuan Cheng and Jinhua Zhu and Jun Ran and Junguang Jiang and Junjie Qiu and Junlong Li and Junxiao Song and Kai Dong and Kaige Gao and Kang Guan and Kexin Huang and Kexing Zhou and Kezhao Huang and Kuai Yu and Lean Wang and Lecong Zhang and Lei Wang and Liang Zhao and Liangsheng Yin and Lihua Guo and Lingxiao Luo and Linwang Ma and Litong Wang and Liyue Zhang and M. S. Di and M. Y Xu and Mingchuan Zhang and Minghua Zhang and Minghui Tang and Mingxu Zhou and Panpan Huang and Peixin Cong and Peiyi Wang and Qiancheng Wang and Qihao Zhu and Qingyang Li and Qinyu Chen and Qiushi Du and Ruiling Xu and Ruiqi Ge and Ruisong Zhang and Ruizhe Pan and Runji Wang and Runqiu Yin and Runxin Xu and Ruomeng Shen and Ruoyu Zhang and S. H. Liu and Shanghao Lu and Shangyan Zhou and Shanhuang Chen and Shaofei Cai and Shaoyuan Chen and Shengding Hu and Shengyu Liu and Shiqiang Hu and Shirong Ma and Shiyu Wang and Shuiping Yu and Shunfeng Zhou and Shuting Pan and Songyang Zhou and Tao Ni and Tao Yun and Tian Pei and Tian Ye and Tianyuan Yue and Wangding Zeng and Wen Liu and Wenfeng Liang and Wenjie Pang and Wenjing Luo and Wenjun Gao and Wentao Zhang and Xi Gao and Xiangwen Wang and Xiao Bi and Xiaodong Liu and Xiaohan Wang and Xiaokang Chen and Xiaokang Zhang and Xiaotao Nie and Xin Cheng and Xin Liu and Xin Xie and Xingchao Liu and Xingkai Yu and Xingyou Li and Xinyu Yang and Xinyuan Li and Xu Chen and Xuecheng Su and Xuehai Pan and Xuheng Lin and Xuwei Fu and Y. Q. Wang and Yang Zhang and Yanhong Xu and Yanru Ma and Yao Li and Yao Li and Yao Zhao and Yaofeng Sun and Yaohui Wang and Yi Qian and Yi Yu and Yichao Zhang and Yifan Ding and Yifan Shi and Yiliang Xiong and Ying He and Ying Zhou and Yinmin Zhong and Yishi Piao and Yisong Wang and Yixiao Chen and Yixuan Tan and Yixuan Wei and Yiyang Ma and Yiyuan Liu and Yonglun Yang and Yongqiang Guo and Yongtong Wu and Yu Wu and Yuan Cheng and Yuan Ou and Yuanfan Xu and Yuduan Wang and Yue Gong and Yuhan Wu and Yuheng Zou and Yukun Li and Yunfan Xiong and Yuxiang Luo and Yuxiang You and Yuxuan Liu and Yuyang Zhou and Z. F. Wu and Z. Z. Ren and Zehua Zhao and Zehui Ren and Zhangli Sha and Zhe Fu and Zhean Xu and Zhenda Xie and Zhengyan Zhang and Zhewen Hao and Zhibin Gou and Zhicheng Ma and Zhigang Yan and Zhihong Shao and Zhixian Huang and Zhiyu Wu and Zhuoshu Li and Zhuping Zhang and Zian Xu and Zihao Wang and Zihui Gu and Zijia Zhu and Zilin Li and Zipeng Zhang and Ziwei Xie and Ziyi Gao and Zizheng Pan and Zongqing Yao and Bei Feng and Hui Li and J. L. Cai and Jiaqi Ni and Lei Xu and Meng Li and Ning Tian and R. J. Chen and R. L. Jin and S. S. Li and Shuang Zhou and Tianyu Sun and X. Q. Li and Xiangyue Jin and Xiaojin Shen and Xiaosha Chen and Xinnan Song and Xinyi Zhou and Y. X. Zhu and Yanping Huang and Yaohui Li and Yi Zheng and Yuchen Zhu and Yunxian Ma and Zhen Huang and Zhipeng Xu and Zhongyu Zhang and Dongjie Ji and Jian Liang and Jianzhong Guo and Jin Chen and Leyi Xia and Miaojun Wang and Mingming Li and Peng Zhang and Ruyi Chen and Shangmian Sun and Shaoqing Wu and Shengfeng Ye and T. Wang and W. L. Xiao and Wei An and Xianzu Wang and Xiaowen Sun and Xiaoxiang Wang and Ying Tang and Yukun Zha and Zekai Zhang and Zhe Ju and Zhen Zhang and Zihua Qu},
      year={2025},
      eprint={2512.02556},
      archivePrefix={arXiv},
      primaryClass={cs.CL},
      url={https://arxiv.org/abs/2512.02556}, 
}

@misc{qwen3technicalreport,
      title={Qwen3 Technical Report}, 
      author={Qwen Team},
      year={2025},
      eprint={2505.09388},
      archivePrefix={arXiv},
      primaryClass={cs.CL},
      url={https://arxiv.org/abs/2505.09388}, 
}

@misc{zhuang2024efficientragefficientretrievermultihop,
      title={EfficientRAG: Efficient Retriever for Multi-Hop Question Answering}, 
      author={Ziyuan Zhuang and Zhiyang Zhang and Sitao Cheng and Fangkai Yang and Jia Liu and Shujian Huang and Qingwei Lin and Saravan Rajmohan and Dongmei Zhang and Qi Zhang},
      year={2024},
      eprint={2408.04259},
      archivePrefix={arXiv},
      primaryClass={cs.CL},
      url={https://arxiv.org/abs/2408.04259}, 
}

@misc{jiapeng2024treereviewstreebaseddynamic,
      title={Tree of Reviews: A Tree-based Dynamic Iterative Retrieval Framework for Multi-hop Question Answering}, 
      author={Li Jiapeng and Liu Runze and Li Yabo and Zhou Tong and Li Mingling and Chen Xiang},
      year={2024},
      eprint={2404.14464},
      archivePrefix={arXiv},
      primaryClass={cs.CL},
      url={https://arxiv.org/abs/2404.14464}, 
}

@article{Jin2024FlashRAGAM,
  title={FlashRAG: A Modular Toolkit for Efficient Retrieval-Augmented Generation Research},
  author={Jiajie Jin and Yutao Zhu and Xinyu Yang and Chenghao Zhang and Zhicheng Dou},
  journal={Companion Proceedings of the ACM on Web Conference 2025},
  year={2024},
  url={https://api.semanticscholar.org/CorpusID:269982691}
}

\appendix

\section{Additional Methods, Experimental Details, and Supplementary Results}
This appendix gathers supporting material moved out of the main paper for space, including fuller method details, supplementary results, and additional analyses that clarify the operating behavior of BEAR.

\subsection{Neither-Case Failure Analysis}
\label{app:neither-failure}
Among the 27 cases where neither path retrieves the supporting evidence, we find 44 missed nodes in total. All missed nodes are leaf nodes, with 93\% occurring at depth 3, and 93\% of the cases come from the medical domain. They are dominated by Multi-hop Reasoning and Comparison queries, indicating that these failures concentrate in settings that require cross-branch evidence collection within structurally homogeneous document types.

\paragraph{LLM-guided path failures.}
The dominant pattern is that the LLM selects the correct document but the wrong tree node (43/44 missed nodes). The main failure mode is a \emph{conclusion-node selection bias}: during in-document node selection, the LLM tends to choose the non-leaf node whose semantics are closest to the query, often a conclusion or summary node. Such a node may summarize the topic correctly, but its subtree does not cover all of the evidence-bearing sibling branches needed for the answer. As a result, when the answer depends on evidence distributed across parallel branches such as chief complaint, physical examination, and auxiliary examinations, selecting a single summary subtree inevitably misses relevant leaf nodes in the other branches. A smaller failure mode appears in multi-document comparison queries, where the LLM sometimes selects the subtree only from the document with the stronger semantic match and misses relevant leaves in the other document.

\paragraph{TreeExpansion failures.}
TreeExpansion misses arise from two sources. The first is \emph{entity-sparse retrieval failure}: when a query is distinguished mainly by proper nouns such as hospital or patient names, generic semantic embeddings may fail to rank the correct document highly enough within a highly homogeneous medical corpus, so the correct document falls outside the top-$k$. The second is \emph{path truncation under the fixed evidence budget}. In some cases, TreeExpansion hits a relevant node in the correct document, but expansion remains confined to one subtree or sibling region, and the 4K token evidence budget is insufficient to cover the deep leaf nodes in other evidence-bearing branches. This is particularly severe in flatter document structures, where multiple relevant branches sit in parallel rather than under a single narrow subtree.

\begin{figure*}[t]
    \centering
    \includegraphics[width=0.95\linewidth]{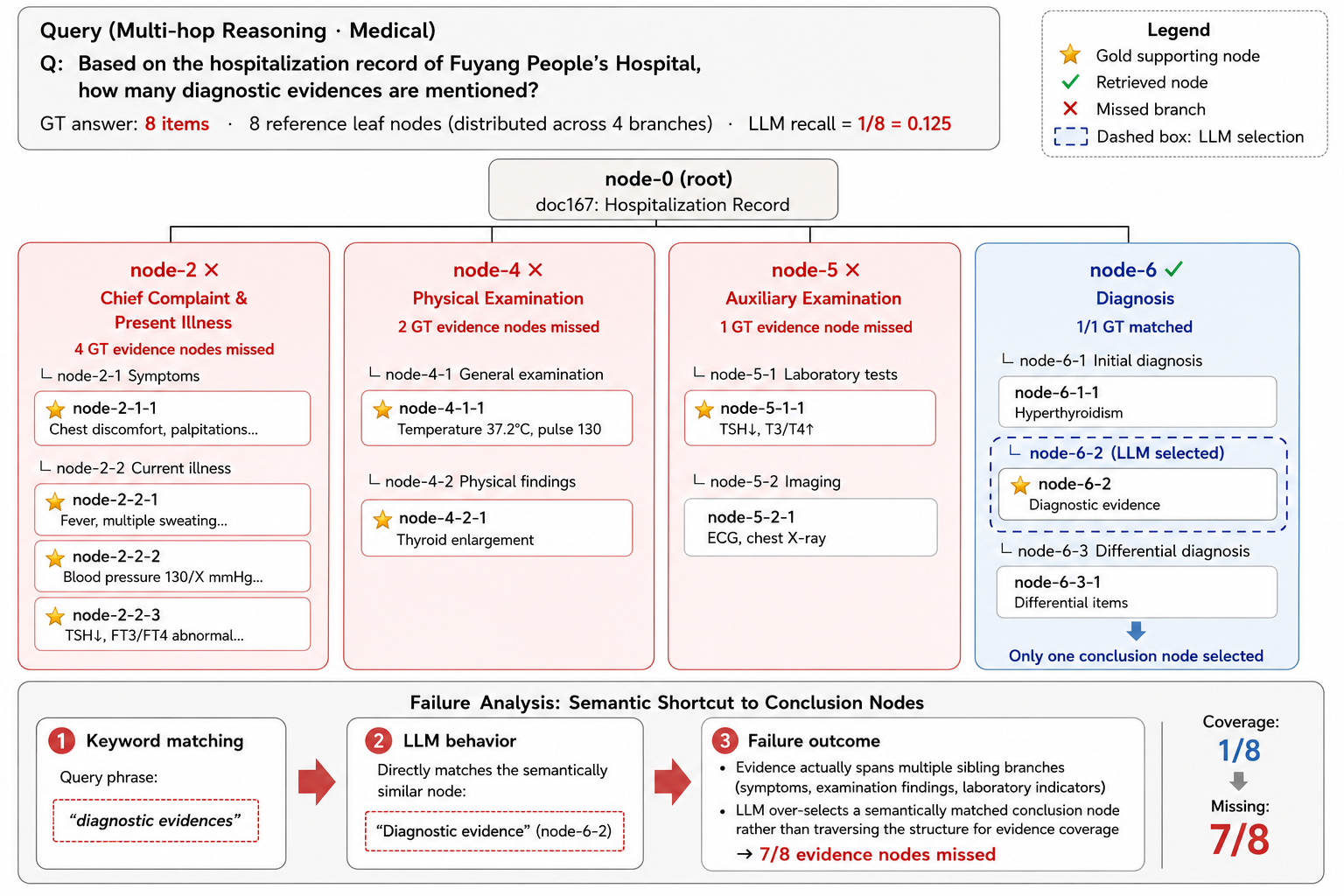}
    \caption{Failure case illustrating semantic shortcut behavior in LLM-guided node selection. The query asks for diagnostic evidence items, but the LLM directly selects a semantically matched conclusion node (``Diagnostic evidence'') instead of retrieving evidence-bearing nodes distributed across multiple sibling branches. As a result, only 1 of 8 gold supporting nodes is covered.}
    \label{fig:failure-case-llm-shortcut}
\end{figure*}

\subsection{Additional Method Details}
\subsubsection{Qwen3-32B Full-Context Control on DragonBall}

Table~\ref{tab:qwen-fullcontext-dragonball} shows that simply exposing the same Qwen3-32B generator to the full document set substantially degrades answer quality relative to structured evidence allocation. Despite receiving more raw context, the full-context control exhibits much lower Completeness and much higher Hallucination and Irrelevance, indicating that increasing available context alone is insufficient. This comparison supports the claim that BEAR's gains arise from selective evidence organization and allocation rather than from the answer model alone.

\subsubsection{Hierarchical Evidence Organization and Multi-Granularity Indexing}
\begin{figure*}[t]
    \centering
    \includegraphics[width=0.9\linewidth]{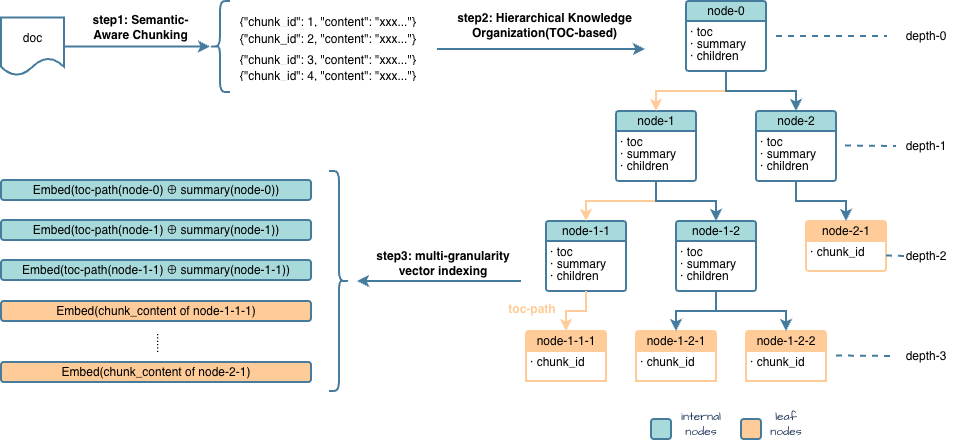}
    \caption{Hierarchical evidence organization and multi-granularity indexing for budgeted evidence access.}
    \label{fig:fable_offline_appendix}
\end{figure*}

This subsection expands the description of BEAR's offline indexing pipeline. Figure~\ref{fig:fable_offline_appendix} summarizes the fuller offline procedure: (1) semantic-aware chunking, (2) LLM-based hierarchical tree generation, and (3) vector indexing of both internal and leaf nodes. For reproducibility, we also include the two core offline prompts used for semantic chunking and tree construction in Appendix Procedure Boxes~\ref{app:prompt1} and~\ref{app:prompt2}, as well as the two query-time prompts used for coarse- and fine-grained LLM-guided evidence selection in Appendix Procedure Boxes~\ref{app:prompt3} and~\ref{app:prompt4}.

\begin{figure*}[t]
\centering
\fbox{\parbox{0.94\textwidth}{
\textbf{Procedure A0.1: Offline prompt for semantic chunking}\par
\small
You are a semantic text chunker.\par
Given the full document TEXT below, your task is to split it into a list of coherent, semantically meaningful chunks.\par
Each chunk should represent one coherent idea or topic, such as a paragraph, a topic segment, or a semantic block.\par
\textbf{Chunking rules:}\par
1. \textit{Semantic coherence:} each chunk should focus on a single topic or idea; if a new topic is introduced, start a new chunk; do not merge content discussing different concepts, even if short.\par
2. \textit{Text preservation:} preserve the original text verbatim; do not paraphrase, summarize, rewrite, or reorder sentences; do not break sentences in half.\par
3. \textit{Coverage and order:} every part of the original text must appear in exactly one chunk; no text may be lost or repeated; chunks must follow the original document order.\par
4. \textit{Length control:} prefer chunks of roughly 100--300 words (or equivalent length); shorter chunks are acceptable if they form a complete semantic unit; avoid overly long chunks unless necessary for semantic completeness.\par
5. \textit{Structure awareness:} if headings or titles appear, group them with the content they introduce; preserve paragraph boundaries and formatting where possible.\par
\textbf{Output format:} each chunk must have exactly two keys: \texttt{"id"} and \texttt{"content"}. Output must be a valid JSON array only, with no explanations, comments, markdown, or code fences.\par
\textbf{Input document:} \texttt{\{document\_text\}}\par
\textbf{Output the chunk list.}
}}
\caption{Offline prompt template used for semantic chunking. We preserve the instruction content while lightly typesetting it for appendix readability.}
\label{app:prompt1}
\end{figure*}

\begin{figure*}[t]
\centering
\fbox{\parbox{0.94\textwidth}{
\textbf{Procedure A0.2: Offline prompt for semantic tree generation}\par
\small
You are a semantic document organizer.\par
You are given a list of pre-segmented chunks of text (\texttt{chunk\_list}), where each chunk has an \texttt{"id"} and \texttt{"content"}. These chunks together represent the full content of a document, split into semantically coherent pieces. The \texttt{"id"} indicates the chunk's original order in the full document.\par
Your task is to analyze the entire set of chunks and produce a hierarchical JSON tree structure that reflects the document's semantic organization.\par
The tree must have the following levels: (1) root, representing the entire document; (2) section, a top-level thematic group; (3) sub\_section, a sub-topic group; and (4) leaf nodes corresponding to the original chunk ids.\par
\textbf{Rules:}\par
1. Every chunk must appear once in the tree; no text may be lost; each chunk can only be assigned to one sub\_section.\par
2. Group chunks based on semantic similarity and topic cohesiveness.\par
3. If a section contains chunks but no distinguishable sub\_sections, group them directly under that section.\par
4. Use natural English labels for section and sub\_section titles that reflect the topics those groups represent.\par
5. Output must be valid JSON with no additional explanation or text.\par
6. Automatically detect the language of the input document and ensure that the output summary is in the same language as the document.\par
\textbf{Output schema:} the JSON tree contains node ids, \texttt{"toc"}, \texttt{"summary"}, and \texttt{"children"}; leaf nodes additionally contain \texttt{"chunk\_id"}.\par
\textbf{Input chunk list:} \texttt{\{chunk\_list\}}\par
\textbf{Every chunk must appear once in the tree. Generate the JSON tree now.}
}}
\caption{Offline prompt template used for hierarchical tree generation from the chunk list.}
\label{app:prompt2}
\end{figure*}

\begin{figure*}[t]
\centering
\fbox{\parbox{0.94\textwidth}{
\textbf{Procedure A0.3: Query-time prompt for document-level LLM-guided selection}\par
\small
You are a document retrieval expert.\par
You are given a user query and a list of documents.\par
Each document is represented by structured tree-based meta information extracted by LLM from the document.\par
Each document meta includes: (1) root-level TOC (document title), (2) root-level summary, and (3) level-1 child sections with TOC + summary when available. These meta fields provide a coarse but semantically rich outline of the entire document. Use this hierarchical structure to judge which documents are semantically relevant to the user's query.\par
\textbf{User Query:} \texttt{\{query\}}\par
\textbf{Available Document Metas:} \texttt{\{doc\_list\_str\}}\par
\textbf{Instructions:}\par
1. Carefully analyze the semantic meaning of the user's query.\par
2. For each document meta, consider all available structural fields, including root TOC, summaries, section titles, level-1 summaries, and any chunk previews.\par
3. Determine how likely the document can help answer the query, based on topic relevance, overlap of key entities/events/actions/dates/organizations, whether the document contains sections matching the query intent, and whether the level-1 sections indicate coverage of the required information.\par
4. Output must be a JSON list of document IDs only, e.g., \texttt{[12, 5, 9]}.\par
5. If no documents are relevant, return an empty list \texttt{[]}.\par
\textbf{Your response (JSON list only):}
}}
\caption{Query-time prompt template used for document-level LLM-guided selection over document summaries and first-level section metadata.}
\label{app:prompt3}
\end{figure*}

\begin{figure*}[t]
\centering
\fbox{\parbox{0.94\textwidth}{
\textbf{Procedure A0.4: Query-time prompt for node-level LLM-guided selection}\par
\small
You are an information retrieval expert.\par
You are given a user query and a list of node descriptions.\par
Each node description is represented by structured tree-based meta information extracted by LLM from the document.\par
Each node meta includes: (1) document id, (2) document title, and (3) structured node information consisting of the TOC + summary fields for the corresponding section level. These meta fields provide a coarse but semantically rich outline of the candidate evidence. Use this hierarchical structure to judge which nodes are semantically relevant to the user's query.\par
\textbf{User Query:} \texttt{\{query\}}\par
\textbf{Available Node Metas:} \texttt{\{nodes\_list\_info\}}\par
\textbf{Instructions:}\par
1. Carefully analyze the semantic meaning of the user's query.\par
2. For each node meta, consider all available structural fields, including document title, TOC path, summaries, section titles, and any chunk previews.\par
3. Determine how likely the node can help answer the query, based on topic relevance, overlap of key entities/events/actions/dates/organizations, whether the node matches the query intent, and whether its local semantic context suggests useful evidence.\par
4. Output must be a JSON list of node IDs only, e.g., \texttt{["170\_node-10", "170\_node-5"]}.\par
5. If no nodes are relevant, return an empty list \texttt{[]}.\par
\textbf{Your response (JSON list only):}
}}
\caption{Query-time prompt template used for node-level LLM-guided selection over structured node metadata.}
\label{app:prompt4}
\end{figure*}

Given a document collection $\mathcal{D} = \{d_1, d_2, \ldots, d_N\}$, BEAR constructs a semantic forest (one tree per document) $\mathcal{F} = \{T_1, T_2, \ldots, T_N\}$, where each tree $T_i$ is a hierarchical abstraction of document $d_i$. In the fuller formulation presented here, each document tree is defined as a rooted tree $T_i=(V_i,E_i)$ with bounded depth and typed nodes (e.g., root, internal summary, and leaf content nodes). Internal nodes capture higher-level semantic abstractions, while leaf nodes retain original semantic chunks.

For each document $d_i$, semantic chunking produces
\begin{equation}
\mathcal{C}_i = \mathrm{LLM}_{\mathrm{segment}}(d_i)=\{c_1,c_2,\ldots,c_{m_i}\},
\end{equation}
where each chunk aligns with a semantically coherent unit rather than a fixed token window. Tree construction then maps the chunk set to a semantic hierarchy:
\begin{equation}
T_i = \mathrm{LLM}_{\mathrm{structure}}(\mathcal{C}_i \mid d_i).
\end{equation}
This process jointly generates a table-of-contents-like hierarchy and node-level summaries. The implementation also supports progressive construction for long documents by building partial trees over partitions and then merging them:
\begin{equation}
\mathcal{T}_i^{(k)} = \textsc{TreeBuild}(\mathrm{part}_k), \quad k=1,\dots,K
\end{equation}
\begin{equation}
\mathcal{T}_i = \textsc{TreeMerge}(\mathcal{T}_i^{(1)}, \dots, \mathcal{T}_i^{(K)}).
\end{equation}

These offline structures provide the substrate for allocating evidence across abstraction levels under a fixed query-time budget. To support multi-granularity retrieval, BEAR embeds non-leaf nodes using their path-level structural context and summary,
\begin{equation}
\mathbf{e}_v = \mathrm{Embed}(\mathrm{toc\_path}(v) \oplus \mathrm{summary}(v)),
\end{equation}
and leaf nodes using the original chunk content,
\begin{equation}
\mathbf{e}_c = \mathrm{Embed}(\mathrm{content}(c)).
\end{equation}
This separation enables both summary-level retrieval and fine-grained evidence retrieval within the same indexed structure. For long documents, the progressive partition-and-merge strategy provides a practical way to scale hierarchy construction beyond a single LLM context window while still preserving document-level semantic coherence after merging the partial trees.

\begin{table*}[t]
\centering
\caption{Default settings used across experiments unless otherwise noted.}
\label{tab:default-settings-appendix}
\small
\begin{tabular}{lcc}
\toprule
Parameter & Default & Note \\
\midrule
Maximum hierarchy depth & $D = 4$ & shared default \\
Retrieval budget & $1\mathrm{K}{-}128\mathrm{K}$ & varies by condition \\
TreeExpansion weights & $(1/3, 1/3, 1/3)$ & default main setting \\
Fixed-length baseline chunk size & 128 & baseline only \\
Chunking / backbone model & dataset-dependent & matches benchmark language \\
Online LLM-guided selection & dataset-dependent & same family as offline structuring \\
Dense retrieval model & BGE-M3 & non-agent experiments \\
FAISS initial recall & top-300 nodes & DragonBall node retrieval \\
Reranked candidates & top-30 nodes & DragonBall node retrieval \\
Reranker threshold & score $> 0.1$ & DragonBall node retrieval \\
Document candidates & top-5 docs & DragonBall node retrieval \\
Node candidates per doc & top-5 nodes & DragonBall node retrieval \\
Retrieval-based generator & Qwen3-32B & shared answer model \\
DeepSeek inference config & provider default & official API setting \\
GPT-OSS / Qwen decoding & model-card default & HuggingFace recommended setup \\
NodeFusion budget rule & ordered truncation & stop at target token budget \\
Budget points in main curves & 1K, 2K, 4K, 8K & main budget sweep \\
\bottomrule
\end{tabular}
\end{table*}

Table~\ref{tab:default-settings-appendix} summarizes the main default settings referenced in Experimental Setup. Across the main-table retrieval-based comparisons, offline hierarchy construction and online LLM-guided document/node selection use DeepSeek-V3.2, which we access through its official API with provider-default settings, and the final answer generator is fixed to Qwen3-32B. LongRefiner is the main exception because its preprocessing relies on a separately trained helper model. To improve reproducibility and reduce implementation variance, we use public released implementations whenever available: BM25 and BGE-M3 in the main results use FlashRAG, while LongRefiner and HippoRAG2 use their official repositories. To keep retrieval comparisons controlled, non-agent pipelines align the dense embedding model to BGE-M3, and retrieval-time LLM steps use DeepSeek-V3.2 unless a baseline does not require them. BrowseComp-plus uses the same DeepSeek-V3.2 retrieval-time configuration while keeping the agent policy model fixed. In the main retrieval-based comparisons, the final evidence context passed to the shared generator is aligned to approximately 4K tokens whenever applicable. When public baseline defaults yield substantially shorter generator-side contexts, we preserve the original retrieval logic and adjust only the final evidence packing or truncation to better match the shared budget. This keeps the comparison focused on retrieval behavior while reducing variation from unequal generator-side context length. In the DragonBall node-retrieval setting used for the main ablation and mechanism analyses, each query first recalls the top 300 nodes from the FAISS index, reranks the top 30, keeps candidates with reranker score above 0.1, selects up to 5 documents after document-level grouping, and then keeps up to 5 node candidates per selected document. Other benchmarks use the same overall retrieval pipeline but may differ in candidate counts and evidence-budget cutoffs depending on corpus scale and task format. At query time, budget control is applied after NodeFusion: we merge explicitly selected LLM nodes and structurally expanded nodes, remove redundant ancestor--descendant overlaps, prioritize LLM-selected nodes, preserve within-document order, and truncate the ordered evidence list once the target evidence budget is reached. We keep this block compact so that it clarifies the shared configuration used throughout the paper without turning the appendix into a full implementation dump.

\subsubsection{Supplementary Retrieval Procedures}
This subsection provides compact procedure summaries for the retrieval pipeline and NodeFusion stage. Together with the offline indexing description above, these procedures show how BEAR first performs document-level budget-aware structured retrieval and only then refines or orders evidence when finer-grained control is needed.

\begin{figure*}[t]
\centering
\fbox{\parbox{0.94\textwidth}{
\textbf{Procedure A1: Extended Budget-Aware Structured Retrieval}\par
\textbf{Input:} Query $q$, Semantic Forest $\mathcal{F}$, hierarchy threshold $L$, budget $B_{\max}$\par
\textbf{Output:} Retrieved content $\mathcal{C}$\par
1. Collect all non-leaf nodes with depth $\leq L$ and expose their $(\mathrm{toc},\mathrm{summary})$ pairs to the LLM.\par
2. Obtain document candidates $\mathcal{D}_{\mathrm{llm}}$ from LLM-guided selection.\par
3. Retrieve dense candidates $\mathcal{D}_{\mathrm{vector}}$ from the node index using FAISS.\par
4. Fuse and deduplicate the two candidate sets to form $\mathcal{D}_{\mathrm{fusion}}$.\par
5. If the total size of the fused document contents is within $B_{\max}$, return them directly.\par
6. Otherwise, perform node-level retrieval with (a) LLM-guided hierarchical navigation and (b) TreeExpansion.\par
7. Merge the selected nodes, order them, and apply budget control to produce the final context.
}}
\caption{Appendix procedure summary for the budget-aware structured retrieval process.}
\label{app:proc1}
\end{figure*}

\begin{figure*}[t]
\centering
\fbox{\parbox{0.94\textwidth}{
\textbf{Procedure A2: NodeFusion}\par
\textbf{Input:} $\mathcal{N}_{\mathrm{llm}}$, $\mathcal{N}_{\mathrm{treexp}}$\par
\textbf{Output:} Ordered chunks $\mathcal{C}_{\mathrm{ordered}}$\par
1. Remove redundant ancestor--descendant overlaps from $\mathcal{N}_{\mathrm{llm}} \cup \mathcal{N}_{\mathrm{treexp}}$.\par
2. Partition the remaining nodes into those directly selected by the LLM and those introduced by TreeExpansion.\par
3. Order documents by LLM-priority first, then append TreeExpansion-only documents.\par
4. Within each document, sort selected nodes by their original position.\par
5. Convert the ordered node sequence to chunks for the generator.
}}
\caption{Appendix procedure summary for the NodeFusion stage.}
\label{app:proc2}
\end{figure*}

These two procedures make the query-time control logic more explicit. Procedure~\ref{app:proc1} shows that BEAR applies the budget check at the document level before invoking finer node-level refinement, which avoids unnecessary processing when coarse evidence already fits the evidence-context budget. Within that node-level stage, TreeExpansion complements direct query--node similarity with ancestor-inherited and child-aggregated structural signals, allowing the retriever to exploit semantic hierarchy rather than relying only on flat embedding-space matching. Procedure~\ref{app:proc2} then shows how NodeFusion performs structure-aware deduplication, priority-based partitioning, and position-preserving ordering so that explicitly selected evidence appears before structurally expanded context while the final prompt remains both compact and readable for the generator.

\begin{table*}[t]
\centering
\caption{Performance on BrowseComp-plus. Top rows are official leaderboard results; the final row replaces the original retriever with BEAR while keeping the agent backbone fixed.}
\label{tab:main-results-agent}
\small
\begin{tabular}{llcccc}
\toprule
LLM & Retriever & Rank & Acc(\%) $\uparrow$ & Recall(\%) $\uparrow$ & SearchCalls $\downarrow$ \\
\midrule
GPT5 & MixedbreadSearch & 1\textsuperscript{st} & \textbf{78.41} & 48.85 & 44.67 \\
GPT5 & Qwen3-Embedding-8B & 3\textsuperscript{rd} & \underline{71.69} & \textbf{78.98} & \textbf{21.74} \\
o3 & Qwen3-Embedding-8B & 4\textsuperscript{th} & 65.90 & 73.24 & 23.97 \\
GPT5 & BM25 & 5\textsuperscript{th} & 57.59 & 61.70 & \underline{23.23} \\
Tongyi-DeepResearch-30B-A3B & Qwen3-Embedding-8B & 11\textsuperscript{th} & 44.46 & 62.32 & 30.37 \\
\midrule
Tongyi-DeepResearch-30B-A3B & \textbf{BEAR(nodes)} & -- & 66.60 & \underline{76.60} & \textbf{21.74} \\
\bottomrule
\end{tabular}
\end{table*}

\begin{table*}[t]
\centering
\caption{Node-level domain-language analysis on DragonBall. The Gain column reports the improvement of bi-path fusion over the stronger single path in each slice. Higher is better.}
\label{tab:domain-language-bipath}
\small
\begin{tabular}{lcccc}
\toprule
Domain-language & LLM-guided (\%) & TreeExpansion (\%) & Bi-path fusion (\%) & Gain \\
\midrule
En-Finance & 87.18 & 94.79 & \textbf{95.35} & +0.56 pp \\
Zh-Finance & 94.42 & 100.00 & \textbf{100.00} & +0.00 pp \\
En-Medical & 91.67 & 86.17 & \textbf{97.62} & +5.95 pp \\
Zh-Medical & 81.17 & 72.26 & \textbf{94.27} & +13.10 pp \\
En-Law & 91.13 & 98.33 & \textbf{99.17} & +0.84 pp \\
Zh-Law & 88.80 & 97.22 & \textbf{99.58} & +2.36 pp \\
\bottomrule
\end{tabular}
\end{table*}

\begin{table*}[t]
\centering
\caption{TreeExpansion weight sensitivity by query type using normalized $(\alpha,\beta,\gamma)$ settings under a 2K budget . The default main-paper setting is $\alpha=\beta=\gamma=\frac{1}{3}$. Higher is better.}
\label{tab:sensitivity}
\small
\begin{tabular}{lcccccc}
\toprule
Query Type & $\frac{1}{3},\frac{1}{3},\frac{1}{3}$ & $0.5,0.25,0.25$ & $0.75,0.25,0$ & $0.25,0.5,0.25$ & $0.25,0.25,0.5$ & $0,0.25,0.75$ \\
\midrule
\multicolumn{7}{l}{\textit{Recall (\%)}} \\
Factual & 94.17 & 93.75 & 93.75 & 92.92 & 95.42 & \textbf{95.42} \\
Info Integration & 74.07 & 76.65 & 79.07 & 71.32 & 77.74 & \textbf{80.03} \\
Temporal Sequence & 75.84 & 80.97 & \textbf{82.22} & 71.11 & 76.95 & 76.39 \\
Multi-hop Reasoning & \textbf{84.01} & 83.68 & 83.68 & 83.48 & 83.42 & 82.92 \\
Overall & 70.93 & 72.44 & \textbf{73.19} & 68.14 & 71.56 & 71.55 \\
\bottomrule
\end{tabular}
\end{table*}

\begin{table*}[t]
\centering
\caption{Progressive node-retrieval ablation by query type on DragonBall. We compare pure leaf-only embedding retrieval, tree-aware embedding retrieval over hierarchical nodes, TreeExpansion, LLM-guided retrieval, and the final bi-path retriever under a 4K budget where applicable. Higher is better.}
\label{tab:progressive-retrieval-ablation}
\small
\begin{tabular}{lccccc}
\toprule
Query Type & Emb (leaf-only) & Emb (Tree) & TreeExpansion & LLM-guided & Bipath (\%) \\
\midrule
Factual & 72.50 & 92.50 & 94.17 & 86.53 & \textbf{99.58} \\
Info Integration & 62.00 & 92.69 & 92.17 & 89.92 & \textbf{99.39} \\
Comparison & 46.11 & 82.64 & 88.82 & 88.47 & \textbf{96.88} \\
Temporal Sequence & 63.96 & 85.28 & 97.22 & 91.94 & \textbf{99.58} \\
Multi-hop Reasoning & 34.35 & 84.22 & 86.32 & 86.67 & \textbf{95.26} \\
Summarization & 24.40 & 90.13 & 90.62 & 91.99 & \textbf{98.47} \\
\midrule
Overall & 50.55 & 87.91 & 91.55 & 89.25 & \textbf{98.19} \\
\bottomrule
\end{tabular}
\end{table*}

\begin{table*}[t]
\centering
\caption{Gold-supporting-evidence attribution by query type at the node level. Bi-path Coverage denotes the union of the two paths, while Neither indicates the fraction of gold-supporting evidence recovered by neither path.}
\label{tab:bipath-overlap}
\small
\begin{tabular}{lccccc}
\toprule
Query Type & Only LLM-guided (\%) & Both (\%) & Only TreeExpansion (\%) & Bi-path (\%) & Neither (\%) \\
\midrule
Factual & 5.3\% & 77.9\% & 16.0\% & 99.2\% & 0.8\% \\
Info Integration & 7.1\% & 82.7\% & 9.0\% & 98.8\% & 1.2\% \\
Comparison  & 7.4\% & 79.1\% & 10.1\% & 96.5\% & 3.5\% \\
Temporal Sequence & 2.5\% & 88.8\% & 8.3\% & 99.6\% & 0.4\% \\
Multi-hop Reasoning & 6.7\% & 72.3\% & 12.6\% & 91.6\% & 8.4\% \\
Summarization & 6.7\% & 82.9\% & 8.6\% & 98.2\% & 1.8\% \\
\midrule
Overall & 6.1\% & 80.7\% & 10.3\% & 97.3\% & 2.9\% \\
\bottomrule
\end{tabular}
\end{table*}

\subsection{Agent Benchmark Details and Results}
BrowseComp-plus tests whether the retrieval gains of BEAR transfer beyond benchmark-style QA into a more realistic agent setting. These tasks require multi-document reasoning, selective evidence acquisition, and iterative access to large document collections, making the benchmark a useful downstream stress test of whether the retriever remains effective when embedded inside a broader reasoning system.

Compared with the official leaderboard entries using the same underlying retriever family, replacing the retriever with BEAR substantially improves both accuracy and recall without changing the agent policy model. This makes the transfer result easier to attribute to retrieval quality rather than to changes in agent reasoning. More broadly, the leaderboard context also shows that agent performance depends jointly on the policy model and the retriever: stronger LLMs tend to help, but retriever quality can still be a major bottleneck. In that sense, the improvement from Tongyi-DeepResearch-30B-A3B/Qwen3-Embedding-8B to Tongyi-DeepResearch-30B-A3B/\textbf{BEAR(nodes)} is informative because it isolates the effect of replacing the retriever while leaving the surrounding agent backbone unchanged. The SearchCalls column is also informative: \textbf{BEAR(nodes)} matches the best reported search-call count in the table while substantially improving over the original Tongyi-DeepResearch-30B-A3B/Qwen3-Embedding-8B system on both accuracy and recall, suggesting that better retrieval quality need not require more agent search steps.

\subsection{Additional Bi-Path Analysis}
Table~\ref{tab:domain-language-bipath} reports the cross-domain and cross-language breakdown. Consistent with the main-text query-type analysis, the node-level bi-path retriever remains stronger than either single path alone across domains and languages. The relative strengths of the two single-path variants vary by slice: the TreeExpansion path is stronger in En-Finance, En-Law, and Zh-Law, whereas the LLM-guided path is stronger in En-Medical and Zh-Medical, with the two paths tied on Zh-Finance. This matches the broader pattern already visible in the query-type analysis: LLM-guided retrieval is more helpful for broad semantic synthesis, whereas the TreeExpansion path is more useful when localized evidence recovery matters, and fusion benefits from combining both signals. The gain over the stronger single path ranges from +0.00 to +13.10 points, with an average improvement of +3.80 points, showing that the complementarity between the LLM-guided path and the TreeExpansion path is not confined to a single topic or language. The largest gains appear in the more challenging Zh-Medical and En-Medical slices, suggesting that fusion is especially helpful when the retriever must combine localized evidence recovery with broader semantic guidance under more difficult conditions.

Table~\ref{tab:progressive-retrieval-ablation} reports a progressive node-retrieval ablation by query type, moving from flat leaf-only embedding retrieval to tree-aware retrieval and then to the stronger single-path and bi-path variants. Appendix Table~\ref{tab:llm-swap-sensitivity} further shows that swapping the offline and online LLMs mainly changes the relative strength of the single-path components, especially the online LLM-guided path, while the final bi-path retriever remains comparatively stable.

The progressive ablation shows a clear staged pattern. The largest gain comes from moving from pure leaf-only embedding retrieval to tree-aware retrieval over hierarchical nodes, indicating that access to internal semantic structure is an important driver of improvement. Building on that tree-aware base, both TreeExpansion and LLM-guided retrieval provide further gains, with their relative strengths varying somewhat by query type. The full bi-path retriever performs best in this appendix comparison, consistent with the complementarity claim in the main text and suggesting that hierarchical indexing and query-time bi-path access are strongest when combined.

\begin{table*}[t]
\centering
\caption{Sensitivity of DragonBall node retrieval to the offline and online LLM choice. We swap the LLM used for offline hierarchy construction and the LLM used for online LLM-guided selection between DeepSeek-V3.2 and GPT-OSS-120B, while keeping the remaining pipeline unchanged. Row names are aligned with the progressive node-retrieval ablation in the main text. Higher is better.}
\label{tab:llm-swap-sensitivity}
\small
\setlength{\tabcolsep}{4pt}
\begin{tabular}{@{}lcccc@{}}
\toprule
Variant &
\shortstack[c]{offline:dsv32\\online:dsv32} &
\shortstack[c]{offline:dsv32\\online:oss120b} &
\shortstack[c]{offline:oss120b\\online:dsv32} &
\shortstack[c]{offline:oss120b\\online:oss120b} \\
\midrule
\multicolumn{5}{l}{\textit{Recall (\%)}} \\
Leaf-only         & 50.55 & 51.11 & 53.64 & 53.64 \\
Tree-aware        & 87.91 & 89.26 & 86.21 & 86.21 \\
Tree + LLM-guided & 89.25 & 71.19 & 79.91 & 69.46 \\
Tree + TreeExp.   & 91.55 & 94.31 & 93.25 & 93.25 \\
\textbf{BEAR(nodes)} & 98.19 & 96.84 & 97.30 & 96.96 \\
\bottomrule
\end{tabular}
\end{table*}

Table~\ref{tab:llm-swap-sensitivity} shows three patterns. First, the original DeepSeek-V3.2/DeepSeek-V3.2 setting remains strongest overall for the final bi-path retriever. Second, the online LLM-guided path is the most sensitive component: replacing the online selector with GPT-OSS-120B substantially reduces the standalone recall of \textit{Tree + LLM-guided}, whereas \textit{Leaf-only}, \textit{Tree-aware}, and \textit{Tree + TreeExp.} are much less affected. Third, the final \textbf{BEAR(nodes)} retriever remains comparatively stable under all four settings, with recall varying only from 96.84\% to 98.19\%. This suggests that the main DragonBall gains are not explained simply by using a stronger LLM, but by the interaction between consistent semantic routing and complementary structural evidence recovery.

\paragraph{Heuristic sensitivity.}
Table~\ref{tab:sensitivity} reports the sensitivity of the TreeExpansion scoring function to different normalized mixing weights over direct similarity, ancestor inheritance, and child aggregation. The default main-paper setting uses equal weights, i.e., $\alpha=\beta=\gamma=1/3$. The query-type breakdown reveals interpretable roles for the three signals. Increasing the direct-similarity weight tends to help multi-document questions such as information integration and temporal sequence, consistent with the need to align semantically related evidence across documents. Stronger child aggregation is most helpful for factual questions, suggesting that bottom-up support from localized descendants is especially useful when evidence is concentrated in specific subtrees. By contrast, larger inheritance weights do not yield comparable gains and can reduce overall recall, indicating that ancestor propagation alone is a weaker discriminative signal than direct matching or child-supported evidence. The balanced default remains a stable compromise even though some query types favor more specialized mixtures.

Table~\ref{tab:bipath-overlap} provides a more direct view of this mechanism by attributing node-level gold-supporting evidence to the LLM-guided path, TreeExpansion, or both across query types.

The table shows that most supported evidence is shared across the two paths, but both also make nonzero unique contributions. TreeExpansion contributes the larger unique share overall (10.3\% vs.\ 6.1\%), while the LLM-guided path still recovers cases that the TreeExpansion path misses. This supports a more precise interpretation of bi-path retrieval: its benefit comes from a strong shared retrieval core plus asymmetric partial complementarity, rather than from two fully redundant or perfectly symmetric mechanisms. Multi-hop Reasoning is the clearest difficult case, with the lowest overlap (72.3\%) and the highest uncovered fraction (8.4\%), which helps explain why combining the two paths is especially valuable when evidence must be assembled across multiple reasoning steps.

\end{document}